\documentclass[letterpaper, 10 pt, journal, twoside]{IEEEtran}

\ifCLASSINFOpdf
\else
\fi

\usepackage{graphicx} 
\usepackage{multirow}
\usepackage{amsfonts}
\usepackage{color}
\usepackage{xcolor}
\definecolor{lightgray}{gray}{0.90}
\usepackage{amsmath}
\usepackage{soul}
\usepackage{soulutf8}
\usepackage{tikz}
\usepackage{wrapfig}
 \usepackage{mathtools}
 \usepackage{nccmath}
\usepackage{caption, subcaption, multicol}
\usepackage{hyperref}

\newcommand{\highlight}[2]{\begingroup\sethlcolor{#1}\textcolor{black}{\hl{#2}}\endgroup}
\newcommand{\capability}{\includegraphics[height=.8em,trim=0 2em 0em 0]{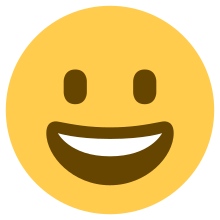}}
\newcommand{\reasoning}{\includegraphics[height=.8em,trim=0 4em 0em 0]{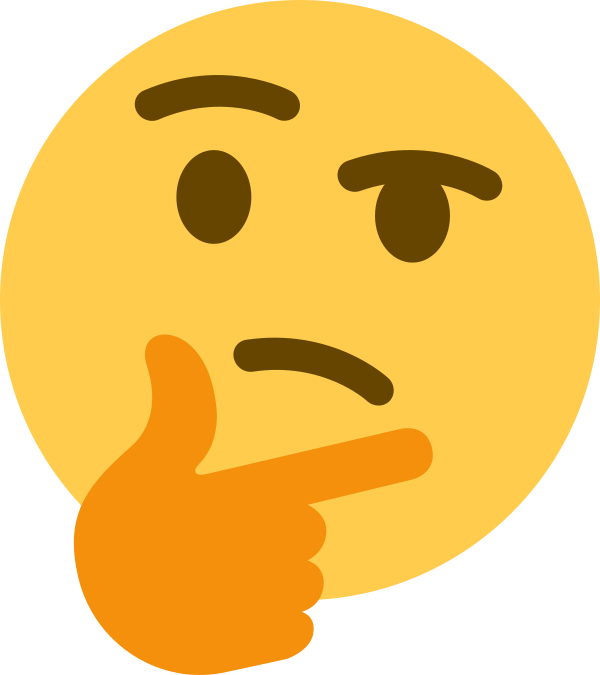}}
\newcommand{\question}{\includegraphics[height=.8em,trim=0 0.7em 0em 0]{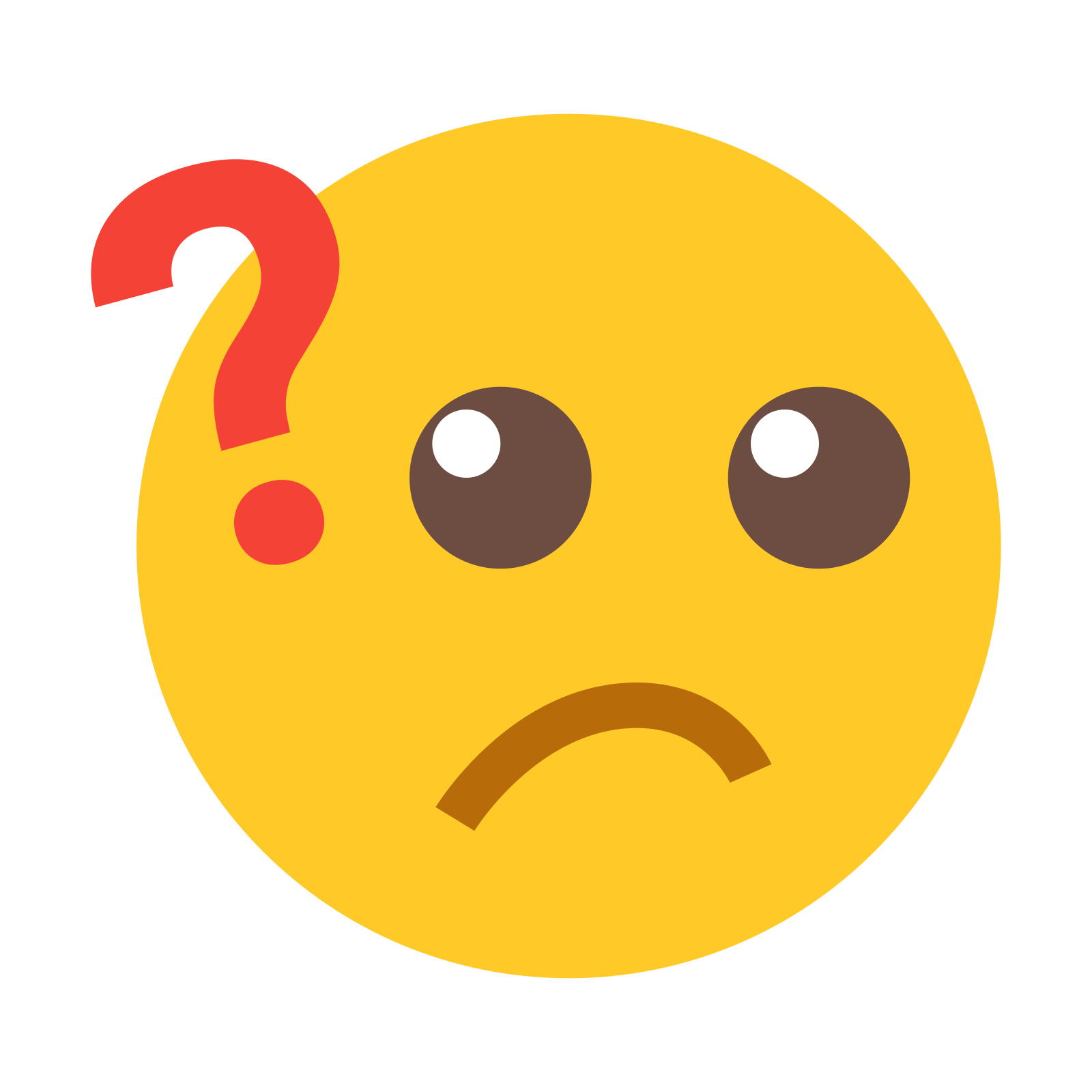}}
\definecolor{bpink}{rgb}{0.99, 0.84, 0.79}
\definecolor{bubbles}{rgb}{0.91, 1.0, 1.0}
\definecolor{champagne}{rgb}{0.97, 0.91, 0.81}

\newcommand\greybo[1]{%
\noindent
\begin{tikzpicture}%
 \node[draw=gray,text width=8.4cm,align=left,
 fill=black!3]
 {\footnotesize{#1}};%
 \end{tikzpicture}%
 }

\newcommand\greybox[1]{%
  \vskip\baselineskip%
  \par\noindent\colorbox{lightgray}{%
    \begin{minipage}{0.48\textwidth} {#1}\end{minipage}%
  }%
  \vskip\baselineskip%
  
}

\newcommand\textbox[1]{%
  \vskip\baselineskip%
  \par\noindent\fbox{%
    \begin{minipage}{0.48\textwidth}{#1}\end{minipage}%
  }%
  \vskip\baselineskip%
}

\newcommand{\blue}[1]{{\color{blue}{\footnotesize \sf{#1}}}}
\newcommand{\red}[1]{{\color{red}{\footnotesize \sf{#1}}}}
\newcommand{\orange}[1]{{\color{orange}\footnotesize {\sf{#1}}}}
\newcommand{\black}[1]{{\color{black} \footnotesize {\sf{#1}}}}
\newcommand{\teal}[1]{{\color{teal}{\sf{#1}}}}
\newcommand{\vio}[1]{{\color{violet}\footnotesize {\sf{#1}}}}
\newcommand{\gray}[1]{{\color{gray}\footnotesize {\sf{#1}}}}
\newcommand{\method}{\textsc{CLARA}}

\title{CLARA: Classifying and Disambiguating User Commands for Reliable Interactive Robotic Agents}

\author{Jeongeun Park$^{1}$,  Seungwon Lim$^{2}$, Joonhyung Lee$^{1}$, Sangbeom Park$^{1}$, \\
Minsuk Chang$^{3}$, Youngjae Yu$^{4}$ and Sungjoon Choi$^{1}$

\thanks{$^{1}$Jeongeun Park, Joonhyung Lee, Sangbeom Park and Sungjoon Choi are with 
the Department of Artificial Intelligence, 
Korea University, Seoul, Korea
{\tt\footnotesize 
\{baro0906, dlwnsgud8823, sangbeom-park, sungjoon-choi\}@korea.ac.kr}.}%
\thanks{$^{2}$ Seungwon Lim is with 
the Department of Computer Science, 
Yonsei University, Seoul, Korea
{\tt\footnotesize \{sngwon\}@yonsei.ac.kr}.}%
\thanks{$^{3}$ Minsuk Chang is with 
Google Research, Seattle, WA, USA
{\tt\footnotesize \{minsukchang\}@google.com}.}%
\thanks{$^{4}$ Youngjae Yu is with 
the Department of Artificial Intelligence, 
Yonsei University, Seoul, Korea
{\tt\footnotesize \{yjy\}@yonsei.ac.kr}.}%
\thanks{Project website: \url{https://clararobot.github.io}}%
\thanks{Digital Object Identifier (DOI): see top of this page.}
}

\begin{document}
\maketitle

\begin{abstract}
    In this paper, we focus on inferring whether the given user command is clear, ambiguous, or infeasible in the context of interactive robotic agents utilizing large language models (LLMs). To tackle this problem, we first present an uncertainty estimation method for LLMs to classify whether the command is certain (i.e., clear) or not (i.e., ambiguous or infeasible). Once the command is classified as uncertain, we further distinguish it between ambiguous or infeasible commands leveraging LLMs with situational aware context prompts. For ambiguous commands, we disambiguate the command by interacting with users via question generation with LLMs. We believe that proper recognition of the given commands could lead to a decrease in malfunction and undesired actions of the robot, enhancing the reliability of interactive robot agents. We present a dataset for robotic situational awareness consisting of pairs of high-level commands, scene descriptions, and labels of command type (i.e., clear, ambiguous, or infeasible). We validate the proposed method on the collected dataset and pick-and-place tabletop simulation environment. Finally, we demonstrate the proposed approach in real-world human-robot interaction experiments.
\end{abstract}

\section{Introduction}
\IEEEPARstart{R}{obotic} agents equipped with large language models (LLMs)~\cite{23_openai,20_brown,23_touvron} have the potential to enhance human-robot interaction by understanding and reasoning about user commands. There is active research \cite{22_huang, 22_ahn, yao2022react} on embracing LLMs into physical robots or utilizing LLMs as planners~\cite{22_huang, 22_ahn} with their reasoning capabilities. On the other hand, as the user's command can be ambiguous or even infeasible, proper interpretation of user commands~\cite{07_brenner} can be an essential component for achieving the reliability of the interactive system. In this paper, we aim to infer whether the given user command is clear, ambiguous, or infeasible in the context of interactive robotic agents utilizing LLMs with situational awareness and interacting with users via question generation on ambiguous scenarios. We propose a framework that can handle both ambiguity and the infeasibility of the language commands while also generating an explanation about the captured uncertainty. 

We aim to classify user commands into clear, ambiguous, and infeasible ones with the awareness of robotic situations and process disambiguation on ambiguous commands. \textit{Situational awareness} is required for this problem, as even the same command can have different meanings for different situations. For instance, when the user command is "he looks tired, can you help him?" with an environment containing coffee, a coffee machine, water, and bread, there can be different interpretations among robot agents. It is certain that the cooking robot, who can only cook food or beverages with a fixed base in the kitchen, should make coffee. Meanwhile, for the cleaning robot, the goal may be categorized as infeasible. This command may be ambiguous for the massage robot, and the robot may ask back to gain further information to disambiguate the lack of information on the command. 



To this end, we propose CLassifying and disAmbiguating user commands for reliable interactive Robotic Agents (\method) to enhance the reliability of the interactive system. Our primary goal is to build an interface that handles the various uncertainties in natural language commands, especially in unstructured raw text (e.g., "he looks sleepy").
The proposed method is composed of two parts: distinguishing a command between clear and not (i.e., ambiguous or infeasible) and classifying ambiguous or infeasible commands for unclear ones. As uncertainty can arise due to both incomplete information or limitations in the agent's capabilities, we first present a method to estimate predictive uncertainty for LLMs. 
Then, we introduce an approach to check feasibility in uncertain commands with situational awareness built upon the zero-shot capability of LLMs to distinguish between infeasible and ambiguous commands. Interacting with users in free-form texts via question generations (i.e., disambiguation) is also conducted on the commands classified as ambiguous. The proposed method can reuse most of the prompts and the structure when the environment changes, with modification on the few-shot demonstrations and the robot capability description. We also designed a benchmark dataset called Situational awareness for Goal Classification in robotic tasks (SaGC), containing pairs of high-level user commands, environments including objects and robot capabilities, and uncertainty types to capture situational awareness in robotic tasks. 

To summarize, the main contributions of this paper are three-fold. 
(1) We introduce a method to capture uncertainty from large language models to recognize ambiguous or infeasible commands.
(2) We propose a technique to classify the type of uncertainty (e.g., ambiguous and infeasible) in the user's command with situational awareness and to track disambiguation progress via free-form text. 
(3) We present a dataset to evaluate the situation-aware uncertainty from large language models consisting of pairs of high-level commands, scene descriptions, and uncertainty-type labels. 

%
%
\section{Related Work}
Building upon the potential of large language models (LLMs), there are several approaches for adapting the reasoning ability of LLMs in robotic planners. 
By incorporating LLMs~\cite{22_huang, 22_ahn,23_liang}, robots can understand and execute tasks described in natural language when confronted with high-level abstract goals to plan for low-level instructions. Utilizing chain-of-thoughts were reported to have success in sequence planning with various source of feedback \cite{22_huang,yao2022react}. There is another approach~\cite{22_ahn, 22_huang_2} to achieve the reliability of LLMs by grounding the generated action to a feasible set. However, interpreting the ambiguous or absurd user command is another element to achieve the reliability of the interactive system. We have a similar approach to Inner Monologue \cite{22_huang}, which gets feedback from the users on ambiguous commands and generates questions with LLMs but differs in that we consider more variety of uncertain commands, e.g., infeasible scenarios.

Interacting with users via question generations to gain additional information for the task is often called disambiguation~\cite{22_yang, 22_pramanick}. Yang et al. \cite{22_yang} proposed Interactive Robotic Grasping with Attribute-Guided Disambiguation by utilizing an attribute-guided POMDP planner for disambiguation. Pramanick et al. \cite{22_pramanick} have proposed an approach for disambiguation by using a Bert-based phrase-to-graph network and deterministic algorithms-based sub-systems. However, it is limited to a fixed set of attributes and template-based question generators. In addition, these approaches can not process free-form text inputs without an additional parser. We believe that approaches without utilizing a template can cover a variety of scenarios with flexibility through natural language generation~\cite{20_brown}.

Uncertainty and ambiguity of the language are actively explored in the question-answering and machine translation domains. There is an approach \cite{21_malinin, 23_kuhn,xiao2019quantifying, 22_xiao,knowno2023} that adapts a tokenwise entropy-based or probability-based approach to estimate the uncertainty. Kuhn et al.~\cite{23_kuhn} leverage the entropy in the meaning space to adapt the semantic equivalence of generated sentences. However, the approaches that require token-wise probability like \cite{21_malinin, 23_kuhn,xiao2019quantifying,22_xiao} can not be applicable to LLMs that can not accessible to those probabilities, e.g., GPT4~\cite{23_openai}.
Fomicheva et al. \cite{20_fomicheva} have explored the uncertainty quantification in machine translation models by exploring the distance in embedding space between sampled generated sentences. 
However, as large language models can have recency bias~\cite{22_lu} with an order of prompts, naive sampling~\cite{23_haung,20_fomicheva} of the outputs may not result in desired diverse samples in uncertain scenarios.

\section{Proposed Method}
\begin{figure*}
    \centering
    \includegraphics[width=0.95\textwidth]{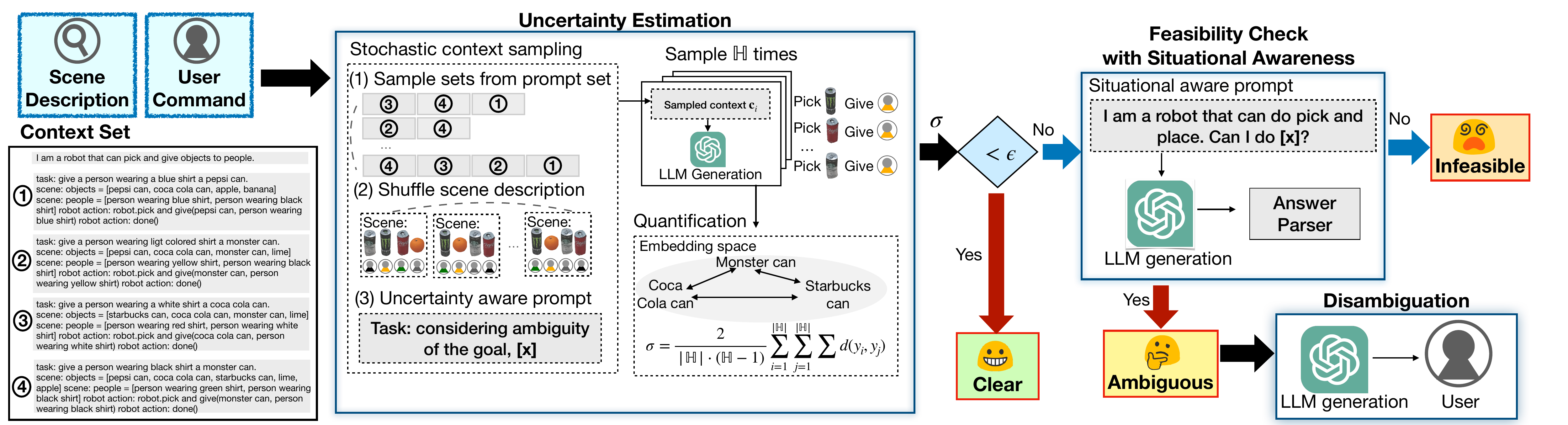}
    \caption{Proposed Method. Our method involves estimating uncertainty with LLMs via context sampling to distinguish between certain and uncertain commands. We then leverage situational awareness to classify uncertain commands into ambiguous and infeasible categories, followed by a disambiguation process for ambiguous commands.
    The number (1) (2), etc., denotes the index of the context from the context set ($C$). $\sigma$ denotes predictive uncertainty, and $\epsilon$ is an uncertainty threshold.}
    \label{fig:method}
\end{figure*}
In this section, we outline our approach to handling uncertain user commands with situational awareness. We begin by presenting a method for estimating uncertainty from large language models. Once uncertainty is estimated, we can distinguish the certain and uncertain inputs with thresholds. We introduce the zero-shot approach to check the feasibility of the uncertain commands, which can classify them into ambiguous and infeasible ones. We present a disambiguation approach to predicted ambiguous inputs in a zero-shot manner. We would like to note that the feasibility check and disambiguation are zero-shot progress, while the uncertainty estimation with the subgoal process has proceeded with few-shot samples.
The proposed method is illustrated in Figure \ref{fig:method}. 

\subsection{Problem Formulation}
In this paper, we focus on capturing and classifying uncertain user commands in the context of interactive agents via large language models (LLMs). 
We assume that the high-level goal command may lack sufficient clarity to execute a task properly, or it may be vague or even an infeasible goal due to users' lack of situational understanding. Our objective is to predict the uncertainty in the LLM's predictions and then predict the type of uncertain goals, i.e., ambiguous or infeasible. If the goal is feasible but ambiguous, we disambiguate the command by generating questions for the user to gather additional information. The input of the system is the high-level goal ($x^g$) and lists of objects in the environment ($\mathbf{x}^s$), along with few-shot contexts ($\mathbf{c}$). LLM then either generates low-level short-horizon skills ($y$), which can be easily interpreted into a robotic action with uncertainty ($\sigma$), or an explanation of the uncertainty via text. 

\subsection{Uncertainty Estimation of LLM}

In this section, we introduce a method to estimate uncertainty on a large language model (LLM). We assume the large language model is operating within an in-context learning framework~\cite{20_brown}. We would like to note that the proposed method does not require additional training and can be utilized even when assessing the model weights or token probabilities is not possible, such as in the case of ChatGPT. 
We conduct context sampling to enforce LLM to generate a more diverse output in uncertain conditions. We hypothesize that the level of certainty in the input would influence the stability and consistency of the LLM's predictions across different contexts~\cite{22_lu}. To enable sampling of the context, we randomly select $k$ contexts from the context set $C$ and shuffle the order of elements within the scene description. As the current LLM is fragile to the recency bias and lacks robustness with a variation of prompts, context sampling can be an effective method to enforce inconsistency in uncertain inputs. 
Our intuition suggests that when a goal is certain, the predictions of LLM will remain relatively consistent, even with slight variations in the context. 
Conversely, when a goal is uncertain, the predictions are expected to exhibit higher variance, contingent upon the specific context provided. The predictive uncertainty $\sigma$ is as follows:

\begin{equation}
    \sigma = \frac{1}{K}\sum_{k}^K\frac{2}{A}\sum_{i}^{\mathbb{H}}\sum_{j}^\mathbb{H} \|(g(\mathbf{y}^k_i) - g(\mathbf{y}^k_j) \|_2
\end{equation}
where $\mathbf{y}_i = f(x^g, \mathbf{x}^s,\mathbf{c}_i)$, $f(\cdot)$ as large language model, $K$ as total number of keywords, and $\mathbf{y}^k_i$ is $k$-th keyword from the generated sentence $\mathbf{y}_i$. The keyword is a phrase that can be manipulated during generation, such as "robot.pick\_and\_give($y_i^1$, $y_i^2$)" in the handover scenario in Figure \ref{fig:method}.
$\mathbb{H}$ is a number of samples, with sampled context $\mathbf{c}_i \sim p(C, \mathbf{x}^s)$, and $A = \frac{2}{\mathbb{H}\cdot (\mathbb{H}-1)}$. During the experiments, we fixed the number of samples $\mathbb{H}$ as $15$. In addition, we define $g$ as a pre-trained mapping function for the word or sentence to a vector space. In particular, we subtract the template (e.g., robot.pick) of generated low-level instruction and leverage word-to-vec embedding \cite{13_mikolov} on the keywords to calculate the pairwise distance of the outputs.

Furthermore, we hypothesize that prompting the large language model to be cognizant of the uncertainty leads to concrete uncertainty quantification. We constructed a prompt at the beginning of the goal input as follows: "\texttt{Considering ambiguity of a goal, [$x^g$]}". This approach, which we refer to as \textit{Uncertainty Aware Prompting}, effectively encourages the language model to acknowledge and account for the uncertainty associated with the given goal. We calculate the empirical Cumulative Distribution Function (CDF) of the uncertainty distribution for a set of samples denoted as "clear", then establish the 80th percentile as the threshold ~\cite{17_richter}.

% ...
For example\footnote{gray denotes prompts, teal as observation, violet denotes generated text from LLMs, and orange denotes signal from a user}, with a set of few-shot prompts $C = \{C_1,C_2,C_3,C_4\}$, and where the task is to "pick a block".
\greybo{\gray{($C_1$) task: pick red block ... ($C_2$) task: pick block colored like a banana ... ($C_4$) ... } \\
Sample 1. \gray{$ \{C_1,C_3, C_4\}$} \\
\gray{task: considering the ambiguity of the goal,} \orange{pick a block} \\
\teal{scene: [red block, yellow block, blue block]} \\
\vio{robot.pick(yellow block)} \\ 
Sample 2. \gray{$ \{C_2,C_4\}$} \\
\gray{task: considering the ambiguity of the goal,} \orange{pick a block} \\
\teal{scene: [red block, yellow block, blue block]} \\
\vio{robot.pick(red block)} \\ 
Sample 3. \gray{$ \{C_3, C_2\}$}\\
\gray{task: considering the ambiguity of the goal,} \orange{pick a block}\\
\teal{scene: [blue block, yellow block, red block]} \\
\vio{robot.pick(blue block)} \\ 
}
{We then sample three generations with different prompts set and shuffled the order of observation. In such case, $\mathbf{y}^1_1,\mathbf{y}^1_2, \mathbf{y}^1_3$ becomes \texttt{"yellow block", "red block", and "blue block"} respectively, with $K=1$ for just one keyword, pick [object]. The distance matrix in word embedding space becomes $\begin{psmallmatrix}0 & 2.5 & 1.7\\ 2.5 & 0 & 2.3 \\ 1.7 & 2.3 & 0\end{psmallmatrix}$, which results in total uncertainty as $2.1$. }

\subsection{Classification and Disambiguation}
As quantifying uncertainty can only determine whether the given command is certain or not, we present a method to analyze and explain the uncertainty. It is composed of three parts: feasibility check, reason generation, and question generation, where all of the procedure is based on prompting LLMs while expanding the previous prompts. Although the proposed method requires few-shot prompts to generate robotic action, we call this classification and disambiguation zero-shot because the prompts provided do not encompass any instances of classification or disambiguation progress. Initially, we perform a feasibility check to assess the viability of the goal in relation to situational awareness by crafting the last line of the prompt with the robot's capability. We first add robot types and the possible actions that the robot can do into a prompt to ensure the agent is aware of their situation; then we force the large language model to conduct a binary classification if a robot can perform the task with answer \texttt{"yes"} or \texttt{"No"}.
With the generated answer, we use a heuristic parser to distinguish commands from infeasible and ambiguous based on the keyword (i.e., \texttt{yes}); if the generated sentence contains the keyword \texttt{"yes"}, we denote the corresponding command as ambiguous.
Continuing from the previous example, the prompt is as follows: 
\greybo{\gray{(Continue from previous prompts)  \\
robot thought: I am a robot that can pick an object. \\
Considering the action set, pick, can I} \orange{pick a block} \gray{? Answer in yes or no\\ answer:}
\vio{
\underline{Yes}, I can pick a block given more information.
}}

If the robot is deemed capable of performing the task, we proceed to disambiguation by generating the reason for the uncertainty and posing a question to the user to gather additional information. 
The reason for the uncertainty and question is generated by prompts like \texttt{"This code is uncertain because"} and \texttt{"What can I ask the user? Please "} to the prompt respectively. 
Again, from the previous example, the prompt for generating an explanation for uncertainty and question are as follows: 

\greybo{\gray{(Continue from previous prompts)  \\
robot thought: this code is uncertain because
}
\vio{the task does not specify any specific criteria for selecting the block}\\
\gray{robot thought: what can I ask to the user?\\
question: Please
}
\vio{provide more information about the criteria for selecting a block 
}
}

After obtaining an answer from the user, the system goes back to the uncertainty estimation step with extended prompts with the disambiguation process.


\section{Experiments}
In our experiments, we aim to address the following research questions with respect to uncertainty with situational awareness: (1) How does the efficacy of our proposed \method, contrast with previous approaches employed for uncertainty quantification across diverse environments?
(2) To what extent can our proposed method accurately identify the user commands that are clear, ambiguous, or infeasible?
(3) What role does the uncertainty-aware interaction module play in clarifying ambiguous commands?
(4) Is it viable to deploy the proposed method in real-world human-robot interaction scenarios? 


\subsection{Baselines}
Regarding uncertainty quantification, we compare the proposed method with four previous approaches: entropy, normalized entropy, semantic uncertainty, and lexical similarity. The predictive entropy, widely recognized as a baseline for uncertainty estimation, is represented as $\mathcal{H} = -\sum_{t=1}^T \sum_{v=1}^V p(y_t^v|x) \log p(y_t^v|x)$, where $V$ is vocabulary size, and $T$ is the sequence length. Normalized entropy \cite{21_malinin} is predictive entropy normalized by sequence length (NE), $\mathcal{H}_{norm} = -\frac{1}{T}\sum_{t=1}^T \sum_{v=1}^V p(y_t^v|x) \log p(y_t^v|x)$.
Semantic entropy (SE) \cite{23_kuhn} estimates the entropy of the random variable representing the output distribution in the semantic event space, as $SE(x) \approx |L|^{-1}\sum_{i=1}^L \log p(L_i|x)$, where semantic equivalence classes are represented as $L_i$. 
Lexical similarity (LS) \cite{20_fomicheva} uses the average similarity of the answers in the answer set, $\mathbb{A}=\frac{2}{|\mathbb{H}|\cdot (|\mathbb{H}|-1)} \sum_{i=1}^{|\mathbb{H}|}\sum_{j=1}^{|\mathbb{H}|} \sum dist(y_i, y_j)$, where the answer set is sampled by beam search of the multinomial distribution.

Furthermore, we validate the effectiveness of the proposed classification and disambiguation method; we compared the proposed method with two different approaches that utilize zero-shot or few-shot capabilities. First is Inner Monologue~\cite{22_huang}, which generates questions and explanations based on the few-shot prompts containing feasibility checks and question generations. In addition, we also compare the proposed method with CLAM\dag~\cite{22_kuhn}, {which requests input that asks if the goal is certain, ambiguous, or infeasible to the large language model} and then further processes disambiguation \footnote{\dag Denotes CLAM~\cite{22_kuhn} modified to zero-shot instead of few-shot.}. We test the method on three different LLMs; LLaMA~\cite{23_touvron}, ChatGPT~(\texttt{GPT3.5-turbo}), and InstructGPT~(\texttt{text-davinci-003}) models via
the OpenAI API\footnote{https://platform.openai.com/docs/model-index-for-researchers}. Inst. GPT denotes InstructGPT in the following tables. In the pick-and-place simulation, we utilize the LLaMA 30B; on the collected dataset, we utilize LLaMA 7B. Due to a lack of computing power, we did not test LLaMA in a real-world environment.

\subsection{Situational Awareness for Goal Classification in Robotic Tasks}
%


In this section, we assess the classification performance with situational awareness of the proposed method. In particular, we aim to evaluate the ability to classify the type of the user's command, e.g., clear, ambiguous, or infeasible, while considering the robotic capabilities and environments. 
We first introduce a dataset specifically designed to evaluate uncertainty in uncertain goals for robotic tasks. 
Then, we measure the method's performance in accurately identifying uncertain commands and categorizing the specific type of uncertainty present in the robot's situation.

\subsubsection{Dataset Formulation}
\begin{figure}[!t]
    \centering
    \includegraphics[width=.48\textwidth]{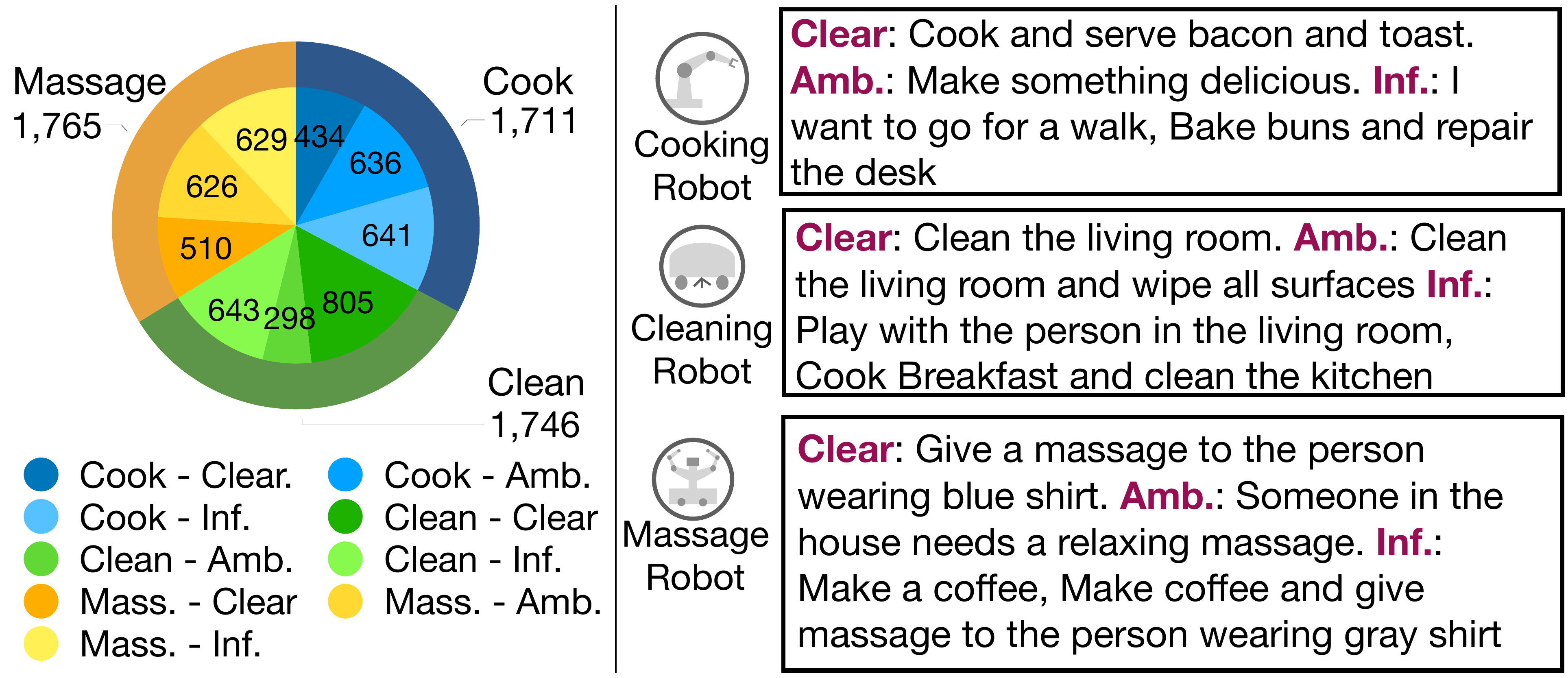}
    \caption{Statistics and Examples of the Dataset. Cer. denotes certain, Inf. denotes infeasible and Amb. denotes ambiguous}
    \label{fig:dataset}
\end{figure}

\begin{table}
    \centering
    \resizebox{0.45\textwidth}{!}{%
    \begin{tabular}{|c|c||c|c|c|c|c|}
    \hline
     &  & LLaMA~\cite{23_touvron} & GPT3.5 & Inst. GPT \\
    \hline 
    \multirow{5}{*}{Quan.}
    & Entropy & 0.714 & - & 0.861 \\
     & NE \cite{21_malinin} & 0.736 & - & 0.867\\
     & SE \cite{23_kuhn} & 0.700 & - & 0.862\\
     & LS \cite{20_fomicheva} & 0.690 & {0.628} & {0.852} \\
    & Ours & {0.725} & {0.710} & {\textbf{0.870}}\\
    \hline  
     \multirow{3}{*}{Cls.} & IM \cite{22_huang} & {0.368} & {0.480} & {0.513}\\
    & CLAM\dag \cite{22_kuhn} & {0.362} & {0.376} & {0.532}\\
    & Ours & {0.447} & {0.556}  & \textbf{{0.710}} \\
    \hline \hline
    \multirow{2}{*}{{Abla.}} &\multirow{2}{*}{Inst. GPT} & {w/o UAP.}  & {w/o CS}. & Ours\\
    \cline{3-5}
    & & {0.861} &  {0.852} & {\textbf{0.870}}\\
    \hline 
    \end{tabular}%
    }
    \caption{Results on SaGC Dataset. Quan. denotes the uncertainty quantification part, where we measured AUROC and Cls. denotes the classification part where we measured accuracy. {Abla. is an ablation study on uncertainty estimation, where CS. denotes context sampling and UAP. denotes uncertainty-aware prompt.}}
    \label{tab:usea}
\end{table}

We first present a dataset called Situational Awareness for Goal Classification in Robotic Tasks (SaGC), curated to evaluate the situation-aware uncertainty of the robotic tasks. 
We collected a dataset consisting of high-level goals paired with scene descriptions annotated with three types of uncertainties inspired by previous work \cite{22_liu, 23_liu} on collecting data via LLM. The primary aim of this dataset is to evaluate whether the language model can effectively distinguish between the three types of goals. 
The dataset consists of 15 different scenes, encompassing three different robot categories: cooking, cleaning, and massaging.
To construct the dataset, we initially created three certain high-level goal examples for each scene and robot, resulting in a total of 105 crafted goals. Crafted examples are formatted into a prompt with the scene description and a type of robot, to be expanded with LLM, i.e., \texttt{gpt-3.5-turbo}. We employed three different prompts to generate goals of varying uncertainty types, which automatically generate the goal based on the given uncertainty label. Furthermore, we also added more complex infeasible commands into the dataset where it is partially feasible, such as \texttt{"bake buns and repair the desk"}. The four validators were then asked to validate the corresponding generated pairs. They were asked to discard the sample, change the label of the sample, or accept the sample. The dataset comprises 5,222 pairs in total, 1,749 certain, 1,560 ambiguous, and 1,917 infeasible goals. The overall statistics and the examples of the dataset are illustrated in Figure \ref{fig:dataset}.


\subsubsection{Results}

We evaluate the proposed method on the dataset by two different measures. First, we measure area under the ROC curve (AUROC) between certain and uncertain high-level goals with the uncertainty quantification baselines. We aim to see how the uncertainty estimation method can separate certain goals from uncertain goals, as shown in Table \ref{tab:usea}. The proposed method with IntructGPT outperforms the compared baselines with a minor gap of $0.003$ and is second-best on LLaMA 7B model. We observe that selecting InstructGPT or GPT3.5 would be a better choice in the proposed method. We would like to emphasize that the proposed method can work compatible with previous methods without the access of token-wise probability. In addition, compared to Lexical Similarity \cite{20_fomicheva}, enforcing the stochastic in the prompts leads to diverse generations in uncertain inputs, as LLM is known to be fragile to recency bias. {Furthermore, we have conducted ablation studies to further analyze the two hypotheses: context sampling and uncertainty-aware prompting. We observe that the performance drops by $0.018$ and $0.09$ without utilizing those modules.}

\begin{figure*}
\centering
    \includegraphics[width=.95\textwidth]{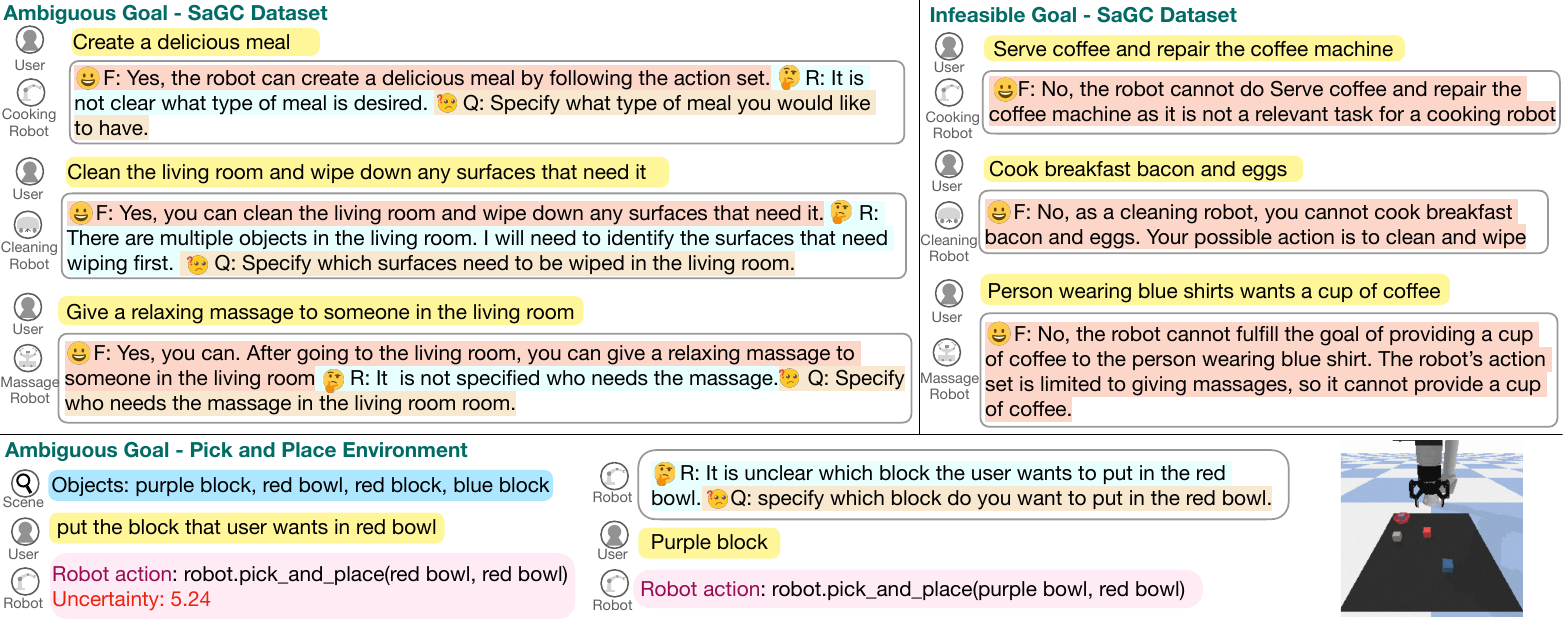}
    \caption{{Examples of generated explanation and question from the proposed method}.  \capability F, \reasoning R, \question Q means Feasibility, Reasoning, and Question, respectively.}
    \label{fig:exp}
\end{figure*}

Furthermore, we measure the classification accuracy of the whole system in Table \ref{tab:usea}, where the proposed method with IntructGPT outperforms the previous method with a gap of $0.158$. We also observe that the proposed method outperforms the baselines in three different large language models. 
We posit that the ability to understand both robotic situations and uncertainty improves accuracy. We observe that a few-shot-based method~\cite{22_huang} can have its weakness in generalization to unseen uncertain commands, and naively asking LLM to conduct a three-way classification has its weakness due to the hallucination issue of LLM and not being fully aware of the situation. We believe that leveraging uncertainty to filter out certain goals and addressing the situational awareness in-context approach afterward improves classification accuracy, making the model easier to predict. 
Although the dataset is formulated by \texttt{gpt-3.5-turbo}, the performance using \texttt{text-danvici-003} (instruct GPT) recorded higher classification accuracy, showing a stronger correlation in the model size of LLMs. 
The illustrations of generated responses are shown in Figure \ref{fig:exp}.


\subsection{Pick and Place Simulations} 

\begin{table}[!t]
    \centering
    \begin{tabular}{|c|c|}
        \hline
         Categories & Tasks\\
        \hline
        \multirow{6}{*}{Clear} & pick [x] and put on [x] bowl\\
        & place all blocks on [x] corner\\
        & place all blocks on [x] bowl \\ 
        & put all blocks on different corners \\ 
        & place blocks on matching color \\
        & place blocks on mismatching color \\
        & stack all blocks on [x] corner \\
        \hline
        \multirow{4}{*}{Ambiguous} & pick block that user wants and place on [x] bowl \\
        & pick [x] block and put on the bowl that the user wants\\
        & pick the block and put in the bowl \\ 
        & stack all blocks \\
        \hline
    \end{tabular}
    \caption{Task explanation in Pick-And-Place Environment}
    \label{tab:pnp_task}
\end{table}

\begin{table}[]
    \centering
    \begin{tabular}{|c|c|c|c|c|c|}
        \hline
         AUORC &  Entropy & NE~\cite{21_malinin} & SE~\cite{23_kuhn} & LS~\cite{20_fomicheva} & Ours\\
         \hline
         LLaMA~\cite{23_touvron}&  0.725 & 0.642 & 0.688 & 0.580 & 0.602\\
         GPT 3.5 & - & - & - & 0.623 & 0.731\\
         Inst. GPT & 0.819 & \underline{0.839} & 0.620 & 0.762 & \textbf{0.864}\\
         \hline \hline
         Ablation & \multicolumn{2}{c|}{Ours w/o UAP.} & \multicolumn{2}{c|}{Ours w/o CS.} & Ours \\
         \hline
         Inst. GPT & \multicolumn{2}{c|}{0.801}  & \multicolumn{2}{c|}{0.762} & \textbf{0.864}\\
         \hline
    \end{tabular}
    \caption{Uncertain Task Detection in Pick-And-Place Environment. CS. denotes context sampling and UAP. denotes uncertainty-aware prompt.}
    \label{tab:pnp_unct}
\end{table}


\begin{table}[!t]
    \centering
    \resizebox{0.40\textwidth}{!}{%
    \begin{tabular}{|c||c|c||c|c||c|c|}
        \hline
          & \multicolumn{2}{c||}{LLaMA~\cite{23_touvron}} & \multicolumn{2}{c||}{GPT3.5} & \multicolumn{2}{c|}{Inst. GPT} \\
         & {F1} & Gap. & {F1} & Gap & {F1} & Gap\\
         \hline
         IM \cite{22_huang}& 0.12 & 0.06 & 0.60& \textbf{0.49} & 0.49 & 0.26\\
         CLAM\dag \cite{22_kuhn} & 0.49 & 0.22 & 0.48 & 0.24 & 0.41 & 0.30\\
         Ours & 0.33 & 0.07 & 0.49 & 0.24 & \textbf{0.64}& 0.39 \\
        \hline
    \end{tabular}%
    }
    \caption{Disambiguation on Pick-and-Place Environment. Gap denotes the success rate gap after interaction on ambiguous commands.}
    \label{tab:pnp_disambiguaton}
\end{table}


In this section, we evaluate the effectiveness of the proposed method on tabletop pick and place simulation. We first aim to observe if the uncertainty estimation method can discriminate certain tasks between ambiguous tasks. In addition, we explore the efficiency of the proposed disambiguation module. We followed the task presented on Inner Monologue (IM) \cite{22_huang}, {which contains eight different types of goals. As these tasks lack the number of ambiguous scenarios}, we added three different types of ambiguous goals and divided certain and ambiguous commands. We define an ambiguous task as a goal that does not specify the exact position or name for either pick or place object, as shown in Table \ref{tab:pnp_task}. We utilize the ViLD \cite{21_gu} for scene description, and Cliport \cite{22_shridhar} for text-to-robotic policy. We evaluated the proposed method on 108 certain configurations and 60 ambiguous configurations, 188 in total. In the interaction phase, we assume that the system can request a question from the user only once.

We first assess the efficacy of our proposed method for uncertainty estimation by measuring the area under the ROC curve (AUROC) between certain and uncertain commands, shown in Table \ref{tab:pnp_unct}. The proposed method with the IntructGPT model outperforms the compared method with margins of $0.025$. In addition, by ablation studies, we observe that the uncertainty-aware prompt showed improvement of the AUROC with a $0.063$ gap, and AUROC increased $0.102$ with context sampling. This leads us to posit that applying both stochastic context and uncertainty-aware prompting enhances the LLM's ability to estimate uncertainty.

Furthermore, we validate the effectiveness of disambiguation in Table \ref{tab:pnp_disambiguaton}. {We have evaluated two main factors in disambiguation progress: the success rate gap after the interaction on ambiguous commands set and the F1 score of the interaction. The F1 score shows the ability of the robot to generate questions only in ambiguous scenarios, where asking for additional information is not necessarily in certain scenarios.
For calculating the F1 score, we divided the commands into two categories: unambiguous (labeled negative) and ambiguous (labeled positive). Although the proposed method had the second-best performance on the success rate gap among baselines, we observe that the proposed method outperforms the previous method in the F1 score metric. We posit that the proposed method generates more questions only when the robot lacks the information while generating appropriate questions to increase the success rate after the interaction.}
The illustration of the interaction with users is shown in Figure \ref{fig:exp}.

\subsection{Real-World Demonstrations}
In this section, we explore the applicability of the proposed method in real-world human-robot handover scenarios. 
We utilize the OWL-VIT \cite{22_minderer} for scene description and grounding translator \cite{22_huang_2} to map the output of LLM to a feasible action set.
Again, We investigate three cases of goal information: clear, ambiguous, and infeasible.
We conduct six different configurations for each label, leading to 18 environments in total. The detailed goals are illustrated in Table \ref{tab:realword_tasks}. 

\begin{table}[t]
    \centering
    \begin{tabular}{|c|c|}
        \hline
         Categories & Tasks\\
        \hline
        Clear & give [x] to [y]\\
        \hline
        \multirow{3}{*}{Ambiguous} & give [x] to someone \\
        & give something to drink to [y] \\
& give something to drink to someone \\ 
        \hline
        \multirow{3}{*}{Infeasible}
        & wipe the desk \\
        & smash the [x] \\
        & put [x] on the ground \\
        \hline
    \end{tabular}
    \caption{Task explanation for the real-world experiment.}
    \label{tab:realword_tasks}
\end{table}

\begin{table}[]
    \centering
    \begin{tabular}{|c|c|c|c|c|c|}
        \hline
         Method & Entropy & NE~\cite{21_malinin} & SE~\cite{23_kuhn} & LS~\cite{20_fomicheva} & Ours\\
         \hline
         GPT3.5 & - & - & - & 0.847 & 0.903\\
         Inst. GPT & 0.958 & 0.972 & 0.847 & 0.951 & \textbf{0.986}\\
         \hline
    \end{tabular}
    \caption{Uncertain Task Detection in Real-World Environment}
    \label{tab:realworld_unc}
\end{table}

\begin{table}[!t]
    \centering
    \resizebox{0.35\textwidth}{!}{%
    \begin{tabular}{|c|c||c|c|c|}
        \hline
         \multirow{2}{*}{Model} & \multirow{2}{*}{Method}& Cls. & \multicolumn{2}{c|}{Disamb.}\\
          & & Acc. &  {F1} & Gap \\
        \hline
         \multirow{3}{*}{GPT3.5} & IM~\cite{22_huang}& 0.44 & 0.36 & 0.17 \\
         & CLAM\dag~\cite{22_kuhn} & 0.44 & 0.67 & 0.28 \\
         & Ours & 0.67 & 0.75 & \textbf{0.33} \\
        \hline
        \multirow{3}{*}{InstructGPT} & IM~\cite{22_huang} & 0.72 & 0.66 & 0.17\\
        & CLAM\dag~\cite{22_kuhn} & 0.55 & 0.61 & 0.06  \\
        & Ours & \textbf{0.89}& \textbf{0.83} & 0.28\\
        \hline
    \end{tabular}%
    }
    \caption{Classification and Disambiguation in Real-World Environment. Cls. denotes uncertainty type classification and Disamb. denotes the disambiguation progress. }
    \label{tab:realworld_int}
\end{table}

\begin{figure*}
\centering
    \includegraphics[width=.98\textwidth]{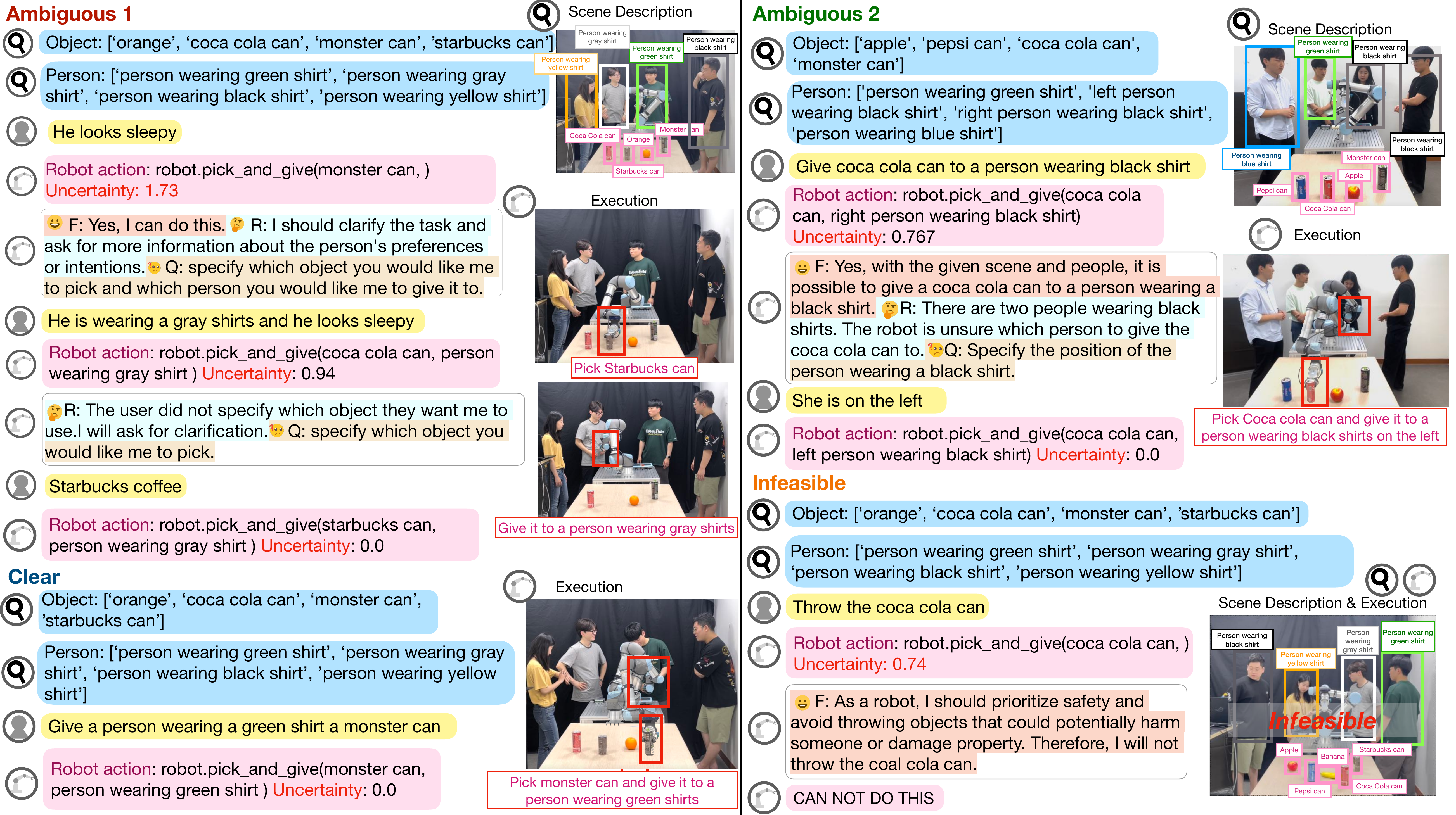}
    \caption{Real-world demonstrations.  \capability F, \reasoning R, \question Q means Feasibility, Reasoning, and Question, respectively.}
    \label{fig:exp_rw}
\end{figure*}

We first measure the AUROC between clear and uncertain commands, i.e., a combination of ambiguous and infeasible, as shown in Table \ref{tab:realworld_unc}. The proposed method with the IntructGPT model outperforms the compared method with a gap of $0.018$. Furthermore, we measure the classification accuracy uncertainty labels in Table \ref{tab:realworld_int}. The proposed method outperforms the classification accuracy on other baselines, with a gap of $0.17$ on both types of LLMs.  
In addition, we measured the success rate gap after disambiguation and F1 score, which is the same metric used in the previous section. The proposed method reported the best success rate increase and appropriate timing for interaction compared to baselines, with an average gap of $0.05$ and $0.23$, respectively. In the real-world environment, the interaction via the GPT3.5-turbo model had a higher success rate than the IntructGPT. {We found that the GPT3.5 model generated more questions both on ambiguous and unambiguous commands, leading to a larger success rate gap. Although asking for user feedback can help increase the success rate gap, a trade-off exists between F1 score measures. Requiring too much user feedback, even on clear commands, may be undesired behavior depending on the user preference~\cite{22_singh}.}

Figure \ref{fig:exp_rw} illustrates the demonstrations of the proposed method in the real-world environment on clear, ambiguous, and infeasible goals using \texttt{gpt-3.5 turbo}. We also tested the system on vague and raw input like "\texttt{he looks sleepy}", with giving only partial information during the first disambiguation progress. We observe that the proposed method successfully understands the raw text inputs and can progress disambiguation iteratively when a user is not giving sufficient information. {We also demonstrated the proposed method under the referential ambiguous scenario, with command \texttt{"give Coca-Cola can to a person wearing a black shirt"}, where two people are wearing black shirts in the scene. The robot explains that two people are wearing black shirts and further asks the user to specify the position (left or right).}
Throughout the demonstrations, we observed that the proposed method could be successfully applied to a real-world environment.

%
%
\section{Limitations}
The proposed method has its limitation of solely relying on the few-shot or zero-shot capability of the LLM, which can be improved via fine-tuning. However, it becomes orthogonal to the contributions we have presented, as the proposed method can be applied without additional models or fine-tuning. 
In addition, as the proposed method is a sampling-based approach, we have limitations on speed and computational cost. Furthermore, calibration to estimate the threshold requires a subset of the clear samples. 
Finally, as the proposed method focuses on uncertainty from the language commands, it has its weakness under a partially observable environment. For example, when the robot needs to find an object that is not yet seen in the environment, the robot regards this planning uncertainty as ambiguous commands and asks for user feedback. For future work, classifying fine-grained uncertainty types (e.g., ambiguity in commands, ambiguity in planning, infeasibility from the environment, or infeasibility from the agent capability) is needed.
For the SaGC dataset introduced in this paper, the dataset may be biased as it is formulated via a large language model. However, we carefully posit that as the LLM used for data construction does not record the best performance, the effect of these biases is less significant. 



\section{Conclusion}
In this paper, we focused on classifying and disambiguating the user commands in the context of interactive robotic agents utilizing large language models (LLMs). In particular, we distinguished the user commands into three different types, i.e., clear, ambiguous, and infeasible. To tackle this problem, we first presented the uncertainty estimation method on LLMs. Then, we introduced an approach to classify the type of uncertain goals (ambiguous or infeasible), and interaction for disambiguation in ambiguous commands. Furthermore, we have presented a dataset to validate the situational awareness of the robotic agent. We evaluated the proposed method on this dataset alongside a pick-and-place simulation environment and real-world demonstration. We observed that the proposed method could properly quantify the uncertainty from LLMs and appropriately classify the type of user commands. We believe that the classification and interaction module can be further developed by fine-tuning methods with data consisting of explanations. 


\bibliographystyle{unsrt}
\bibliography{reference}

\newpage
\appendix

\subsection{Failure Cases} 

\label{sec:failure}
\paragraph{Threshold} The uncertainty, threshold is $1.24$ in the cooking robot case, where the uncertainty is $1.22$ which is slightly lower than the threshold. For the case under, the system classifies ambiguous commands into certain commands. 
\textbox{
\black{Goal: Cook something delicious for me.} \\
\black{Robot Type: Cooking robot} \\
\blue{GT label: Ambiguous} \\
\vio{robot action: robot.grab(bacon)} \\
\red{Uncertainty: 0.59} \\
\vio{robot action: robot.heat(bacon)} \\
\red{Uncertainty: 0.0}\\
\vio{robot action: robot.grab(bread)} \\
\red{Uncertainty: 0.82}\\
\vio{robot action: robot.heat(bread} \\
\red{Uncertainty: 0.0}\\
\vio{robot action: robot.plate(bacon)} \\ 
\red{Uncertainty: 1.22} \\
\black{=============} \\
\blue{Pred label: Certain}
}
\paragraph{Known keywords in the infeasible command} If there is a known word in the goal in the infeasible commands, the large language models fail to predict it as an infeasible command. For example, in the case of a cooking robot, if the goal input contains words related to cooking, such as "after cooking a meal", the LLM predicts its goal as feasible. 
\textbox{
\black{Goal: Clean the kitchen after cooking a meal with bacon and banana} \\
\black{Robot Type: Cooking robot} \\
\blue{GT label: infeasible} \\
\teal{....} \\
\vio{robot action: robot.plate(bacon)} \\
\red{Uncertainty: 1.49} \\
\highlight{bpink}{\capability \small \sf F:  Yes, the robot can clean the kitchen after cooking a meal with bacon and banana. The robot can grab the banana and bacon, heat them, plate them, and then clean the kitchen by grabbing the water, desk, and table and wiping them down.} \\
\black{=============} \\
\blue{Pred label: Ambiguous}
}
In addition, even though the possible action set is "clean" and "wipe" for the cleaning robot, in the "Organize the desk" commands, the LLM predicts that "organize" is in the action set. This sufficient understanding of the action set leads to failure, which can be relaxed via more prompts or fine-tuning approaches. 
\textbox{
\black{Goal: Organize the desk} \\
\black{Robot Type: Ceaning robot} \\
\blue{GT label: infeasible} \\
\vio{robot action: robot.clean(desk)} \\
\red{Uncertainty: 0.93} \\
\highlight{bpink}{\capability \small \sf F: Yes, the robot can organize the desk. The robot action would be robot.organize(desk)} \\
\black{=============} \\
\blue{Pred label: Ambiguous}
}
\paragraph{Ambiguous commands with missing keywords} Furthermore, the proposed methods sometimes fail on too vague commands with missing words. For example, "give a massage in the kitchen" can be interpreted as "give a massage to a person in the kitchen". However, the proposed method predicts it is infeasible because of the missing word "person" in this case. 
\textbox{
\black{Goal: Give a massage in the kitchen} \\
\black{Robot Type: Massage robot} \\
\blue{GT label: Ambiguous} \\
\vio{robot action: robot.go to(kitchen)} \\
\red{Uncertainty: 1.17} \\
\highlight{bpink}{\capability \small \sf F: No, the robot cannot give a massage in the kitchen as it is not part of its action set. The robot can only "go to" and "give massage" to the person wearing the yellow shirt, white shirt, black shirt, or brown shirt.} \\
\black{=============} \\
\blue{Pred label: infeasible}
}

\paragraph{Lack of Information in Question}
In the scenario where the system has multiple missing information, the LLM fails to generate a question asking for both of the information. 
\textbox{
\black{Goal: Give someone something to drink} \\
\black{Objects = [coca cola can, fanta can, lemon, apple]} \\
\blue{People = [person wearing black shirts, person wearing white shirts]} \\
\vio{robot.pick\_and\_give(coca cola can, person wearing blue shirt)} \\
\red{Uncertainty: 1.75} \\
\highlight{bpink}{\capability \small \sf F: Yes, I can give someone something to drink.} \\
\highlight{bubbles}{\reasoning \small \sf R: It is not specified which person should receive the drink.} \\
\highlight{champagne}{\question \small \sf Q: Specify which person should receive the drink} \\
}

\begin{figure*}[!t]
    \centering
    \includegraphics[width=0.99\textwidth]{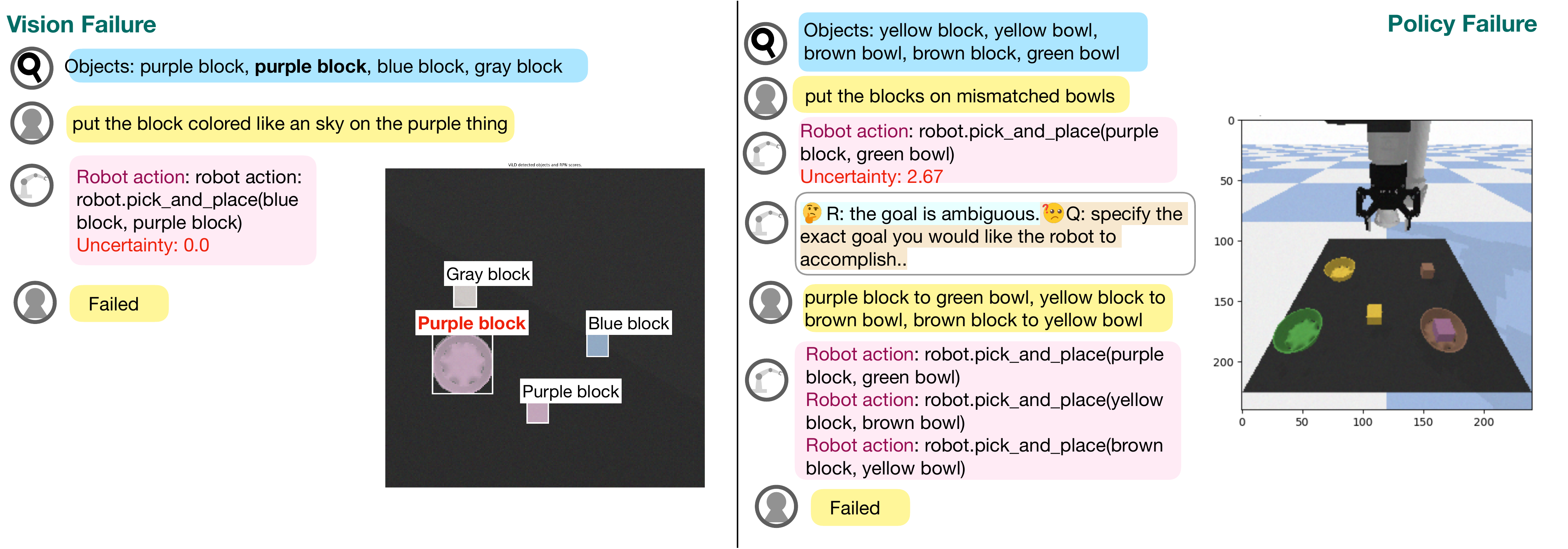}
    \caption{Failure Cases due to other modules}
    \label{fig:failure}
\end{figure*}

\paragraph{Malfunction in other modules} Malfunctions on the vision module or robotic policy module can lead to wrong uncertainty quantification or overall failure. The examples are shown in Figure \ref{fig:failure}. 

\subsection{Toy Example}
Toy's example of a simple block-picking scenario is in Section III-B. In this example, the context sets $C = \{C_1,C_2,C_3,C_4\}$ are prompted as follows: \\
\greybox{\gray{($C_1$) task: pick a red block -
scene: [red block, blue block, yellow block] -
robot action: robot.pick(red block)\\
($C_2$) task: pick a block colored like a banana - scene: [green block, blue block, yellow block] - robot action: robot.pick(yellow block) \\
($C_3$) task: pick something that resembles a tree - scene: [green block, blue block, yellow block, red block] - robot action: robot.pick(green block) \\
($C_4$) task: pick something that is not blue - scene: [blue block, purple block] - robot action: robot.pick(purple block)} }
where the goal is \texttt{"pick a block"} with observing \texttt{red block, blue block, yellow block} in the scene. We first sample the contexts from $C$ (e.g., $\{C_1,C_3, C_4\}$) and shuffle the scene description list. Then, $\mathbf{c_1}$ will be as follows: \\
\greybox{\gray{($C_1$) task: pick a red block -
scene: [red block, blue block, yellow block] -
robot action: robot.pick(red block)\\
($C_3$) task: pick something that resembles a tree - scene: [green block, blue block, yellow block, red block] - robot action: robot.pick(green block) \\
($C_4$) task: pick something that is not blue - scene: [blue block, purple block] - robot action: robot.pick(purple block)}\\
\gray{task: considering the ambiguity of the goal,} \orange{pick a block} \\
\teal{scene: [red block, yellow block, blue block]} }

Similarly, if we sample $ \{C_3,C_2\}$ and shuffle the scene description list, $\mathbf{c_3}$ will be as follows:\\
\greybox{\gray{($C_3$) task: pick something that resembles a tree - scene: [green block, blue block, yellow block, red block] - robot action: robot.pick(green block) \\
($C_2$) task: pick block colored like a banana - scene: [green block, blue block, yellow block] - robot action: robot.pick(yellow block)
\gray{task: considering the ambiguity of the goal,} \orange{pick a block} \\
\teal{scene: [blue block, yellow block, red block]} }
}

We then sample three generations with different prompts set and shuffled the order of observation.

\greybox{Sample 1. \gray{$\{C_!,C_3, C_4\}$} \\
\gray{task: considering the ambiguity of the goal,} \orange{pick a block} \teal{scene: [red block, yellow block, blue block]} \\
\vio{robot.pick(yellow block)} \\ 
Sample 2. \gray{$\{C_2,C_4\}$} \\
\gray{task: considering the ambiguity of the goal,} \orange{pick a block} \teal{scene: [red block, yellow block, blue block]} \\
\vio{robot.pick(red block)} \\ 
Sample 3. \gray{$\{C_3, C_2\}$}\\
\gray{task: considering the ambiguity of the goal,} \orange{pick a block} \teal{scene: [blue block, yellow block, red block]} \\
\vio{robot.pick(blue block)} \\ 
}
In such case, $\mathbf{y}_1,\mathbf{y}_2, \mathbf{y}_3$ becomes \texttt{"yellow block", "red block", and "blue block"} respectively. 
The distance matrix in word embedding space will be $\begin{psmallmatrix}0 & 2.5 & 1.7\\ 2.5 & 0 & 2.3 \\ 1.7 & 2.3 & 0\end{psmallmatrix}$, which results in total uncertainty as $2.1$. 

In the classification section, the prompts and the generated outcome of the sample example are as follows: 

\greybox{\gray{(Continue from previous prompts)  \\
robot thought: I am a robot that can pick an object. \\
Considering the action set, pick, can I} \orange{pick a block} \gray{? answer in yes or no} \\
\vio{
\underline{Yes}, I can pick a block given more information.
}}
With the generated answer, we used a heuristic parser to distinguish commands from infeasible and ambiguous based on the keyword (i.e., \texttt{yes}). If the generated sentence contains the keyword \texttt{"yes"}, we denote the corresponding command as ambiguous. In the toy example, the command is classified as ambiguous. 

For the disambiguation progress, the prompts and generated outcome are as follows:

\greybox{\gray{(Continue from previous prompts)  \\
robot thought: this code is uncertain because
}
\vio{the task does not specify any specific criteria for selecting the block}\\
\gray{(robot thought: what can I ask to the user?
question: Please
}
\vio{provide more information about the criteria for selecting a block
}
}
The system provides the reason for the uncertainty and the following question for disambiguation. 
 After obtaining the user's answer, the system returns to the uncertainty estimation step with extended prompts with the disambiguation process.

\greybox{Sample 1. \gray{
($C_4$), ($C_1$) task: considering the ambiguity of the goal,} \orange{pick a block} \teal{scene: [red block, yellow block, blue block]}\\
\vio{(Append Generated Explanation and Question)} \\
\gray{user: }
\orange{red block}\\
\vio{robot.pick(red block)}
}

\subsection{Demonstration in Mobile Manipulation Task}
To analyze the proposed method in the larger environment under partial observability, we have demonstrated the proposed method in Alfworld~\cite{20_shridhar}, following the prompt baseline from ~\cite{yao2022react}. The threshold for uncertainty is set as $1.5$ in this environment. 

In certain cases, the uncertainty remains lower than the threshold. 
\textbox{
\black{
You are in the middle of a room. Looking quickly around you, you see a cabinet 6, a cabinet 5, a cabinet 4, a cabinet 3, a cabinet 2, a cabinet 1, a coffeemachine 1, a countertop 3, a countertop 2, a countertop 1, a drawer 3, a drawer 2, a drawer 1, a fridge 1, a garbagecan 1, a microwave 1, a shelf 3, a shelf 2, a shelf 1, a sinkbasin 1, a stoveburner 4, a stoveburner 3, a stoveburner 2, a stoveburner 1, and a toaster 1.\\
Your task is to:} \orange{put a saltshaker in drawer 1.} \\
\vio{Act 1: go to countertop 1} \\
\red{Uncertainty 1: 1.21}\\
\black{Obs 1: On the countertop 1, you see a creditcard 2, a dishsponge 2, and a mug 2.}\\
\vio{Act 2: go to countertop 2}\\
\red{Uncertainty 2: 0.0}\\
\black{Obs 2: On the countertop 2, you see a creditcard 1, a cup 2, a knife 2, a knife 1, a lettuce 1, a papertowelroll 1, a soapbottle 2, a statue 1, a tomato 2, and a tomato 1.} \\
\vio{Act 3: go to countertop 3}\\
\red{Uncertainty 3: 0.0}\\
\black{Obs 3: On the countertop 3, you see a bread 1, a butterknife 1, a houseplant 1, a potato 1, a spoon 1, and a vase 2.} \\
\vio{Act 4: go to shelf 1} \\
\red{Uncertainty 4: 0.0} \\
\black{Obs 4: On the shelf 1, you see a mug 3, and a statue 2.} \\
\vio{Act 5: go to shelf 2} \\
\red{Uncertainty 5: 0.0} }
\textbox{
\black{Obs 5: On the shelf 2, you see a creditcard 3, and a dishsponge 3.}\\
\vio{Act 6: go to shelf 3}\\
\red{Uncertainty 6: 0.0}\\
\black{Obs 6: On the shelf 3, you see a vase 1.} \\
\vio{Act 7: go to cabinet 1} \\
\red{Uncertainty 7: 0.713} \\
\black{Obs 7: On the cabinet 1, you see a glassbottle 1.}
\vio{Act 8: go to cabinet 2} \\
\red{Uncertainty 8: 0.0} \\
\black{Obs 8: The cabinet 2 is closed.}\\
\vio{Act 9: open cabinet 2}\\
\black{Uncertainty 9: 0.0} \\
\black{Obs 9: You open the cabinet 2. The cabinet 2 is open. In it, you see a plate 1, a saltshaker 1, and a soapbottle 1.} \\
\vio{Act 10: take saltshaker 1 from cabinet 2} \\
\red{Uncertainty 10: 0.0} \\
\black{Obs 10: You pick up the saltshaker 1 from the cabinet 2.} \\
\vio{Act 11: go to drawer 1} \\
\red{Uncertainty 11: 0.0}\\
\black{Obs 11: The drawer 1 is closed.} \\
\vio{Act 12: open drawer 1}\\
\red{Uncertainty 12: 0.0}\\
\black{Obs 12: You open the drawer 1. The drawer 1 is open. In it, you see a fork 1, and a peppershaker 1.} \\
\vio{Act 13: put saltshaker 1 in/on drawer 1}\\
\red{Uncertainty 13: 0.0}\\
\black{Obs 13: You put the saltshaker 1 in/on the drawer 1.}
\black{=============} \\
\blue{Pred label: certain}
}

In the ambiguous scenario, where the goal is to \texttt{put something in the drawer 1}, the uncertainty incerases in the beginning, asking the user to clarify what something is. During proceeding with the task, the robot feels that it is stuck, and asks the user where the \texttt{saltshaker} is. 

\textbox{
\black{You are in the middle of a room. Looking quickly around you, you see a cabinet 6, a cabinet 5, a cabinet 4, a cabinet 3, a cabinet 2, a cabinet 1, a coffeemachine 1, a countertop 3, a countertop 2, a countertop 1, a drawer 3, a drawer 2, a drawer 1, a fridge 1, a garbagecan 1, a microwave 1, a shelf 3, a shelf 2, a shelf 1, a sinkbasin 1, a stoveburner 4, a stoveburner 3, a stoveburner 2, a stoveburner 1, and a toaster 1.\\
Your task is to:} \orange{put something in the drawer 1} \\
\vio{Act 1: go to cabinet 1} \\
\red{Uncertainty 1: 1.55} \\
\highlight{bpink}{\capability \small \sf F: Yes.} 
\highlight{bubbles}{\reasoning \small \sf R: First I need to find something I can put in the drawer 1.} \\
\highlight{champagne}{\question \small \sf Q: Tell me what item you would like to put in the drawer 1.} \\
\orange{User: saltshaker}\\
\vio{Act 2: go to countertop 1}\\
\red{Uncertainty 2: 0.647}}
\textbox{
\black{Obs 2: On the countertop 1, you see a creditcard 2, a dishsponge 2, and a mug 2.}\\
\vio{Act 3: go to countertop 2}\\
\red{Uncertainty 3: 2.14}\\
\highlight{bubbles}{\reasoning \small \sf R: The saltshaker is not here.} \\
\highlight{champagne}{\question \small \sf Q: Tell me where the saltshaker is located.} \\
\orange{User: cabinent 2} \\
\vio{Act 4: go to cabinet 2}\\
\red{Uncertainty 4: 0.0}\\
\black{Obs 4: The cabinet 2 is closed.} \\
\vio{Act 5: open cabinet 2}\\
\red{Uncertainty 5: 0.0}\\
\black{Obs 5: You open the cabinet 2. The cabinet 2 is open. In it, you see a plate 1, a saltshaker 1, and a soapbottle 1.} \\
\vio{Act 6: take saltshaker 1 from cabinet 2} \\
\red{Uncertainty 6: 0.0} \\
\black{Obs 6: You pick up the saltshaker 1 from the cabinet 2.} \\
\vio{Act 7: go to drawer 1} \\
\red{Uncertainty 7: 0.0} \\
\black{Obs 7: The drawer 1 is closed.} \\
\vio{Act 8: open drawer 1} \\
\red{Uncertainty 8: 0.0} \\
\black{Obs 8: You open the drawer 1. The drawer 1 is open. In it, you see a fork 1, and a peppershaker 1.} \\
\vio{Act 9: put saltshaker 1 in/on drawer 1}\\
\red{Uncertainty 9: 0.0} \\
\black{Obs 9: You put the saltshaker 1 in/on the drawer 1.} \\
\black{=============} \\
\blue{Pred label: ambiguous}
}

In the infeasible scenario, the uncertainty becomes high in the beginning, and tells the user that the command is infeasible. 
\textbox{
\black{You are in the middle of a room. Looking quickly around you, you see a cabinet 6, a cabinet 5, a cabinet 4, a cabinet 3, a cabinet 2, a cabinet 1, a coffeemachine 1, a countertop 3, a countertop 2, a countertop 1, a drawer 3, a drawer 2, a drawer 1, a fridge 1, a garbagecan 1, a microwave 1, a shelf 3, a shelf 2, a shelf 1, a sinkbasin 1, a stoveburner 4, a stoveburner 3, a stoveburner 2, a stoveburner 1, and a toaster 1. \\
Your task is to:} \orange{do laundry}\\
\vio{Act 1: go to cabinet 1}\\
\red{Uncertinaty 1: 1.538}\\
\highlight{bpink}{\capability \small \sf F: No.} \\
\black{=============} \\
\blue{Pred label: infeasible}
}

\subsection{SaGC Data collections}
The template to generate the dataset is as follows.
\greybox{
\black{I am a [x] robot. Your possible action set is [y].  Here are the scene and the examples of the goal that [x] robot can do. }\\
\black{objects = [objects here]}\\
\black{floorplan = [floorplan here]} \\ 
\black{people = [people here]} \\ 
\black{Examples here} \\
\blue{(Clear goal)} \black{create the high-level clear goal similar to the examples} \\
\blue{(Ambiguous goal)} \black{create the creative goal that lacks the information to do the task but still can do.} \\
\blue{(infeasible goal)} 
\black{create the simple goal that only [x] robot cannot do}
}
An example prompt for cooking the robot in scene nine is as follows. 
\greybox{
\black{I am a cooking robot. Your possible action is grab, heat and plate. Here are the scene and the examples of the goal that robot can do. }\\
\black{objects = [coffee, table, bread, desk, buns, water, bacon, pan, banana, apple]}\\
\black{floorplan = [kitchen, office, meeting room, gym]} \\ 
\black{people = [person wearing gray shirt", person wearing white shirt, person wearing black shirt, person wearing brown shirt]} \\ 
\black{make breakfast consisted of bacon, bread, and coffee} \\
\black{I just worked out help me} \\
\black{cook a meal with apple and bread} \\
\blue{(Clear goal)} \black{create the high-level clear goal similar to the examples} \\
\blue{(Ambiguous goal)} \black{create the creative goal that lacks the information to do the task but still can do.} \\
\blue{(Infeasible goal)} 
\black{create the simple goal that only cooking robot cannot do}
}
 The annotator is a large language model, and four validators were not blind to the research question. For annotation, we crafted 105 samples and then utilized \texttt{gpt-3.5-turbo} to generate the 5226 pairs of commands and the corresponding label. Utilizing LLMs to construct datasets is efficient yet powerful, which has been widely used in prior studies~\cite {hagrid,23_inpars}. 
The four validators were then asked to validate the corresponding generated pairs. The validators were given three options: discard the sample, change the label of the sample, or accept the sample. We provided guidelines to the validators, specifying that the term \texttt{ambiguous} pertains to scenarios where multiple answers could be applicable or when ambiguous keywords such as \texttt{or, any} are present. For the \texttt{infeasible scenario}, we instructed that it refers to situations either irrelevant to the type of robot or concerning the presence of an object in the scene. We have asked the validator to discard the sample if they can not decide. We observe that approximately $10\%$ of the label changed after the validation. The dataset may have a bias as we have utilized only \texttt{gpt-3.5-turbo} model to generate all the pairs, while the effect of validators is not significant. 

\subsection{SaGC Data Examples}
After validation, the SaGC dataset is composed as follows. Label 0, 1, 2 is certain, ambiguous, and infeasible respectively. 
\greybox{
\sf \small
"scene": \{\\
    "floorplan": [
        "kitchen",
        "living room",
        "bedroom"
    ],\\
    "objects": [
        "water",
        "bacon",
        "bread",
        "pan",
        "coffee",
        "table",
        "orange juice",
        "sasuage",
        "banana",
        "apple"
    ],\\
    "people": [
        "person wearing blue shirt",
        "person wearing white shirt",
        "person wearing red shirt"
    ]\\
\},\\
"goal": "Prepare a meal consisting of bacon, toast and coffee and serve it on the table.",\\
"label": 0,\\
"task": "cooking"\\
===================================
\\
"scene": \{\\
    "floorplan": [
        "kitchen",
        "living room",
        "bedroom",
        "bathroom"
    ],\\
    "objects": [
        "bacon",
        "bread",
        "pan",
        "coffee",
        "table",
        "orange juice",
        "sasuage"
    ],\\
    "people": [
        "person wearing yellow shirt",
        "person wearing white shirt",
        "person wearing black shirt"
    ]\\
\},\\
"goal": "Clean the living room and wipe the table in the kitchen.",\\
"label": 0,\\
===================================
\\
"task": "cleaning"\\
"scene": \{\\
    "floorplan": [
        "kitchen",
        "living room",
        "bedroom"
    ],\\
    "objects": [
        "water",
        "bacon",
        "bread",
        "pan",
        "coffee",
        "table",
        "orange juice",
        "sasuage",
        "banana",
        "apple"
    ],\\
    "people": [
        "person wearing blue shirt",
        "person wearing white shirt",
        "person wearing red shirt"
    ]\\
\},\\
"goal": "Prepare and serve a meal or drink for the user.",\\
"label": 1,\\
"task": "cooking"\\
===================================
\\
"scene": \{\\
    "floorplan": [
        "kitchen",
        "living room",
        "bedroom"
    ],\\
    "objects": [
        "water",
        "bacon",
        "bread",
        "pan",
        "coffee",
        "table",
        "orange juice",
        "sasuage",
        "banana",
        "apple"
    ],\\
    "people": [
        "person wearing blue shirt",
        "person wearing white shirt",
        "person wearing red shirt"
    ]\\
\},
"goal": "Person needs relaxation",\\
"label": 1,\\
"task": "massaging"\\
===================================
}
\greybox{
\sf \small
"scene": \{\\
    "floorplan": [
        "kitchen",
        "office",
        "desk for students",
        "workspace for robots"
    ],\\
    "objects": [
        "coffee",
        "table",
        "bread",
        "desk",
        "robot",
        "orange juice",
        "lemon",
        "salad"
    ],\\
    "people": [
        "person wearing yellow shirt",
        "person wearing white shirt",
        "person wearing black shirt",
        "person wearing blue shirt"
    ]\\
\},\\
"goal": "Teach a person wearing a yellow shirt how to cook a meal.",\\
"label": 2,\\
"task": "cleaning"\\
===================================\\
"scene": \{\\
    "floorplan": [
        "bedroom",
        "living room",
        "kitchen",
        "bathroom"
    ],\\
    "objects": [
        "coffee",
        "table",
        "bread",
        "desk",
        "buns",
        "water",
        "bacon",
        "pan",
        "banana",
        "apple"
    ],\\
    "people": [
        "person wearing gray shirt",
        "person wearing white shirt",
        "person wearing black shirt",
        "person wearing blue shirt"
    ]\\
\},\\
"goal": "person wearing black shirt wants a glass of water",\\
"label": 2,\\
"task": "massaging"\\
}



\subsection{Supervised Learning Baselines}
We have conducted additional experiments on the supervised classification model (i.e., BERT~\cite{19_devlin}), as shown in Table \ref{tab:sl}. First, we trained the BERT with 1000, 500, and 300 train samples and evaluated the rest of the samples of the dataset (a total of 5222 pairs). We observe that with more than 500 samples, the supervised learning method can exceed the proposed method. However, using the supervised learning method can be vulnerable when the configuration of the robot changes. To see the generalizability, we trained the BERT on two different robot types and then tested on the pairs from the test robot type. In this robot-type transfer scenario, the average accuracy was $0.333$, which shows the significant gap of $0.367$ compared to the proposed method on \texttt{text-danvici-003}.

\begin{table}[!h]
    \centering
    \begin{tabular}{|c|c|c|c|c|}
        \hline
        & \multicolumn{2}{c|}{Method} & Acc. \\
        \hline
      \multirow{3}{*}{SL}  
        & \multicolumn{2}{c|}{1000-shots} & 0.926 \\
        & \multicolumn{2}{c|}{500-shots} & 0.716 \\
        & \multicolumn{2}{c|}{300-shots}& 0.521 \\
        \hline
        \multirow{5}{*}{Transfer} & Train & Test & Acc.\\
        \cline{2-4}
        & Clean + Cook & MAS. & 0.455\\
        & Cook + MAS.& CLEAN & 0.298\\
        & Clean + MAS. & COOK  & 0.245\\
         \cline{2-4}
        & \multicolumn{2}{c|}{Average} & 0.333 \\
        \hline
        Zero-shot & \multicolumn{2}{c|}{Ours (\texttt{text-danvici-003})} & 0.710 \\
        \hline
    \end{tabular}
    \caption{Supervised Learning Baselines}
    \label{tab:sl}
\end{table}

\subsection{Vision Language Model}
\begin{figure}[!h]
    \centering
\includegraphics[width=.45\textwidth]{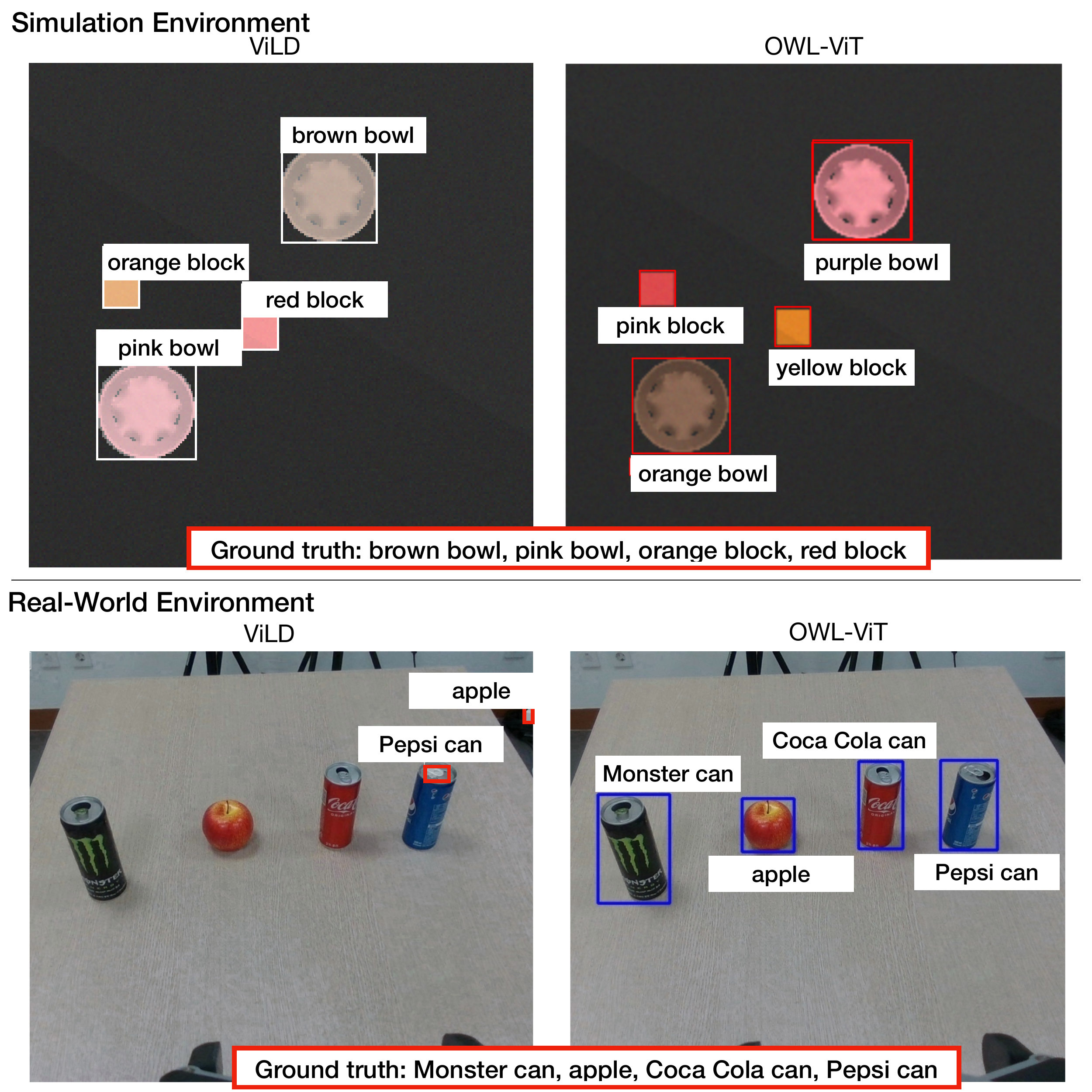}
    \caption{Comparision in Vision-Language Models}
    \label{fig:vision}
\end{figure}

 We selected vision-language models that excel in the environment. In the pick-and-place simulation setting, constructed similarly to SayCan~\cite{22_ahn}, we discovered that using ViLD~\cite{21_gu}, as done in the original paper, outperforms the use of OWL-ViT~\cite{22_minderer}. However, the trend reverses in real-world scenarios. Figure \ref{fig:vision} provides a comparison, illustrating the distinctions between the two vision-language models.

\subsection{Pick and Place Simulation}
\paragraph{Environmnet} We conduct experiments on the table-top pick and place environment with the Pybullet simulator. Following the previous approach on language model-based robots \cite{22_huang}, we utilize ViLD \cite{21_gu} as a vision-language detection model to extract the scene information, a large language model to estimate low-level goals, and Cliport \cite{22_shridhar} to transverse textual goals to robotic action. In the interaction phase, we assume that the system can request a question from the user only once. We define an ambiguous task as a goal, not specifying the pick or place object, as shown in Table \ref{tab:pnp_task}.

\begin{table}[!h]
    \centering
    \begin{tabular}{|c|c|}
        \hline
         Categories & Tasks\\
        \hline
        \multirow{6}{*}{Clear} & pick [x] and put on [x] bowl\\
        & place all blocks on [x] corner\\
        & place all blocks on [x] bowl \\ 
        & put all blocks on different corners \\ 
        & place blocks on matching color \\
        & place blocks on mismatching color \\
        & stack all blocks on [x] corner \\
        \hline
        \multirow{4}{*}{Ambiguous} & pick block that user wants and place on [x] bowl \\
        & pick [x] block and put on the bowl that the user wants\\
        & pick the block and put in the bowl \\ 
        & stack all blocks \\
        \hline
    \end{tabular}
    \caption{Task explanation. We define an ambiguous task as a goal, not specifying the pick or place object.}
    \label{tab:pnp_task}
\end{table}

\paragraph{F1 Measure}
F1 measure shows the ability of the robot to generate questions only in ambiguous scenarios, as asking for additional information is not necessary for clear and infeasible scenarios. For calculating the F1 score, we divided the commands into two categories: unambiguous (labeled negative) and ambiguous (labeled positive), as shown in Table \ref{tab:confusion}. We have added a statement about the importance of the F1 metric in the revised manuscript.

\begin{table}[!h]
    \centering
    \begin{tabular}{|c c|c|c|}
        \hline
        &  & \multicolumn{2}{c|}{Predicted} \\
        & & Negative & Positive \\
        \hline
        \multirow{4}{*}{Actual}& \multirow{2}{*}{Negative} & skip question in & give question in\\
        & & unambiguous command & ambiguous command\\
        \cline{2-4}
        & \multirow{2}{*}{Positive} & skip question & give question in \\
        & & unambiguous command & ambiguous command\\
        \hline
    \end{tabular}
    \caption{Confusion Matrix for calculating F1 score}
    \label{tab:confusion}
\end{table}

\paragraph{Detailed Results} For the details of disambiguation evaluation in Table \ref{tab:pnp_details}, we report the success rate for both before and after an interaction, with the portion of question generation. 
\begin{table*}[!h]
    \centering
    \begin{tabular}{|c|c|c|c|c|c|c|c|c|c|c|c|}
    \hline
         \multirow{2}{*}{Model} & \multirow{2}{*}{Method} & \multicolumn{4}{c|}{Clear} & \multicolumn{4}{c|}{Ambiguous} \\
         & & Before & After & Gap & Por. & Before & After & gap & Por. \\
         \hline
         \multirow{3}{*}{LLAMA~\cite{23_touvron}} & IM \cite{22_huang} & 32.4 & 34.3 & 2.1 & 0.05 & 33.3 & 38.9 & 5.6 & 0.07\\
         & CLAM\dag \cite{22_kuhn}& 25.9 & 40.7 & 14.8 & 0.98 & 27.8 & 50.0 & 22.2 & 0.93\\
         & Ours & 28.7 & 31.4& 2.7 & 0.39 & 22.2 & 29.4& 7.2 & 0.35 \\
         \hline         
         \multirow{3}{*}{GPT3.5} & IM \cite{22_huang} & 36.1 & 40.7 & 4.6 & 0.46& 12.5 & 61.1 & \textbf{48.9} & 0.81 \\
         & CLAM\dag \cite{22_kuhn} & 30.5 & 39.8 & 9.3 & 0.81 & 22.2 & 46.2 & 24.0 & 0.81 \\
         & Ours & 32.4 & 38.4& \underline{6.0}& 0.32 & 20.3 & 44.4 & 23.9 & 0.50 \\
         \hline
         \multirow{3}{*}{IntructGPT} & IM \cite{22_huang} & 30.6 & 34.3 & 3.7 & 0.88 & 22.2 & 48.1& 25.9 & 0.88\\
         & CLAM\dag \cite{22_kuhn} & 30.5 & 25.5 & -5.0 & 0.37 &  42.6 & 13.0 & 29.6 & 0.44 \\
         & Ours & 38.3 & 39.8 &  1.5  & 0.27 & 16.6 & 55.6 & \underline{39.0} & 0.72 \\
         \hline
    \end{tabular}
    \caption{Success Rate with Portion of question generation}
    \label{tab:pnp_details}
\end{table*}

\subsection{GPT-Info Evaluation Examples}
\begin{table}[!h]
    \centering
    \begin{tabular}{|c|c|c|c|}
    \hline
        & LLAMA~\cite{23_touvron}& GPT3.5 & IntructGPT \\
        \hline
        IM \cite{22_huang} &  0.70 & 0.72 & 0.79\\
        CLAM\dag & 0.37 & 0.83 & 0.82\\
        Ours & 0.39 & 0.93 & 0.89 \\
        \hline
    \end{tabular}
    \caption{Informativeness of generated questions and reasons}
    \label{tab:pnp_info}
\end{table}
We assessed the informativeness of the generated questions and explanations with a fine-tuned GPT from TrustfulAI~\cite{22_lin}.
We measure how many questions and reasons generated are informative with respect to the total number of generations. 
The results are shown in Table \ref{tab:pnp_info}. We observe that the proposed method with GPT3.5-turbo generates the most informative questions, and the zero-shot based method (Ours, CLAM\dag \cite{23_kuhn}) had better quality in average than the few-shot based method (IM \cite{22_huang}). We believe that leveraging zero-shot performance in  question generation contributes to better quality and variety in the resulting questions, as opposed to a one-shot approach. 

\greybox{
\black{goal: hide purple block} \\ 
\teal{question: provide more information about the goal of the task.} \\ 
\blue{info: false}\\ 
\black{=======}\\
\black{goal: put the blocks in their matching bowls} \\ 
\teal{question: specify which bowl the blue block should go in.} \\ 
\blue{info: true}\\ 
\black{=======}\\
\black{goal: put the blocks on mismatched bowls} \\ 
\teal{question: specify the exact goal you would like the robot to accomplish.} \\ 
\blue{info: false}\\ 
\black{=======}\\
\black{goal: put the blocks on mismatched bowls} \\ 
\teal{question: provide more information about the task. } \\ 
\blue{info: false}\\ 
\black{=======}
\black{goal: put all the blocks in different corners} \\ 
\teal{question: specify which corner you would like the red block to be placed in.} \\ 
\blue{info: true}\\ 
\black{=======}}
\greybox{
\black{goal: put a blue block in a bowl that user wants. } \\ 
\teal{question: specify which bowl you would like the blue block to be placed in.} \\ 
\blue{info: true}\\ 
\black{=======}
}

Examples of the evaluation by GPT-info from Truthful QA \cite{22_lin} in are shown Box A. 

\subsection{Prompts}

\paragraph{SaGC Dataset} Prompts for cooking robots are as follows. 
\greybox{
\sf \small
I am a cooking robot. Your possible action is grab, heat and plate. \\
objects: [bread, pan, table, water] \\
goal: cook me toast \\
robot action: robot.grab(bread) \\
robot action: robot.heat(bread) \\
robot action: robot.plate(bread)  \\
robot action: robot.done()\\

objects: [table, watter, coffemachine] \\
goal: give a person wearing black shirt a coffee \\
robot action: robot.grab(coffee) \\
robot action: robot.pour(coffee)  \\
robot action: robot.done()\\

objects: [bread, pan, bacon, watter, table] \\
goal: give me a sandwich \\
robot action: robot.grab(bread) \\
robot action: robot.heat(bread) \\
robot action: robot.grab(bacon) \\
robot action: robot.heat(bacon) \\
robot action: robot.plate(bread) \\
robot action: robot.plate(bacon) \\
robot action: robot.done()}
\greybox{
\sf \small

objects: [water, table, pan, bread]\\
goal: I am thirsty\\
robot action: robot.grab(water)\\
robot action: robot.pour(water)\\
robot action: robot.done()\\
From this, predict the next action by considering the role of the robot and the ambiguity of the goal \\
objects: [\orange{Objects here}]\\
goals: \orange{Goal Here}
}
Prompts for cleaning robots are as follows. 
\greybox{
\sf \small
I am a cleaning robot. Your possible action is clean and wipe. \\
scene: [kitchen, living room, bedroom, bathroom] \\
objects: [table, pan, bread] \\
goal: clean the kitchen\\
robot action: robot.clean(kitchen)\\
robot action: robot.done()\\

scene: [kitchen, living room, bedroom]\\
objects: [table, pan, bread] \\
goal: clean bedroom\\
robot action: robot.clean(bedroom)\\
robot action: robot.done()\\

scene: [kitchen, living room, bedroom] \\
objects: [table, pan, bread] \\
goal: clean table\\
robot action: robot.wipe(table)\\
robot action: robot.done()\\

scene: [kitchen, living room, bedroom] \\
objects: [table, pan, desk, bed] \\
goal: clean desk \\
robot action: robot.wipe(desk) \\
robot action: robot.done() \\

From this, predict the next action by considering the role of the robot and the ambiguity of the goal \\
scene: [\orange{Scene here}]\\
objects: [\orange{Objects here}]\\
goals: \orange{Goal Here}
}
Prompts for massage robots are as follows. 
\greybox{
\sf \small
I am a massage robot. Your possible action is go\_to and give\_massage. \\

people: [person wearing black shirt, person wearing yellow shirt, person wearing red shirt] \\
goal: Give a massage to a person wearing black shirt\\
robot action: robot.go\_to(person wearing black shirt)\\
robot action: robot.give\_massage(person wearing black shirt)\\
robot action: robot.done()\\

poeple: [person wearing green shirt, person wearing green shirt, person wearing blue shirt]\\
goal: Give a massage to a person wearing green shirt\\
robot action: robot.go\_to(person wearing green shirt)\\
robot action: robot.give\_massage(person wearing green shirt)\\
robot action: robot.done()\\

poeple: [person wearing green shirt, person wearing green shirt, person wearing blue shirt]\\
goal: A person wearing green shirt needs a massage\\
robot action: robot.go\_to(person wearing green shirt)\\
robot action: robot.give\_massage(person wearing green shirt)\\
robot action: robot.done()\\

people: [person wearing black shirt, person wearing yellow shirt, person wearing red shirt] \\
goal: A person wearing yellow shirt needs a massage\\
robot action: robot.go\_to(person wearing yellow shirt)\\
robot action: robot.give\_massage(person wearing yellow shirt)\\
robot action: robot.done()\\

From this, predict the next action by considering the role of the robot and the ambiguity of the goal \\
people: [\orange{people here}] \\
goals: \orange{Goal Here} \\
}
Prompts for classification and disambiguation is as follows. This is appended to the previous prompts.
\greybox{
\sf \small
Question: I am a [\orange{robot type here}] robot. Considering the action set, Can I [\orange{goal here}]? \\
Answer: \highlight{bpink}{\capability Generated Feasibility}
\\
Robot thought: this code is uncertain because \highlight{bubbles}{\reasoning Generated Reason} \\
Robot thought: what can I ask to user? \\
Question: Please \highlight{champagne}{\question Generated Question} \\
Answer: \orange{User's feedback} \\
robot thought: continue the previous task based on the question and answer
}

\paragraph{Pick and Place}The prompts for the pick and place environment are as follows. 

\greybox{\sf \small task: move all the blocks to the top left corner.\\
scene: objects = [red block, yellow block, blue block, green bowl] \\
robot action: robot.pick\_and\_place(blue block, top left corner) \\
robot action: robot.pick\_and\_place(red block, top left corner) \\
robot action: robot.pick\_and\_place(yellow block, top left corner) \\
robot action: done() }
\greybox{
\sf \small

task: put the yellow one the green thing.\\
scene: objects = [red block, yellow block, blue block, green bowl]\\
robot action: robot.pick\_and\_place(yellow block, green bowl)\\
robot action: done()\\

task: move the light colored block to the middle.\\
scene: objects = [yellow block, blue block, red block] \\
robot action: robot.pick\_and\_place(yellow block, middle) \\
robot action: done() \\

task: stack all the blocks. \\
scene: objects = [blue block, green bowl, red block, yellow bowl, green block] \\
robot action: robot.pick\_and\_place(green block, blue block) \\
robot action: robot.pick\_and\_place(red block, green block) \\
robot action: done() \\
task: group the blue objects together. \\
scene: objects = [red block, blue block, green bowl, blue bowl, yellow block, green block] \\
robot action: robot.pick\_and\_place(blue block, blue bowl) \\
robot action: done() \\

task: put all blocks in the green bowl. \\
scene: objects = [red block, blue block, green bowl, blue bowl, yellow block] \\
robot action: robot.pick\_and\_place(red block, green bowl) \\
robot action: robot.pick\_and\_place(blue block, green bowl) \\ 
robot action: robot.pick\_and\_place(yellow block, green bowl) \\
robot action: done()\\

task: considering the ambiguity of the goal, \orange{Goal Here} where the place object is not always dependent on the selected pick object \\
scene: objects = [\orange{Detection Results}] 
}

Prompts for disambiguation are as follows. This is appended to the previous prompts.
\greybox{ \sf \small
Robot thought: this code is uncertain because \highlight{bubbles}{\reasoning Generated Reason} \\
Robot thought: what can I ask to user? \\
Question: Please \highlight{champagne}{\question Generated Question} \\
Answer: \orange{User's feedback} \\
robot thought: continue the previous task based on the question and answer
}
\paragraph{Real-world Environment} The few-shot contexts are as follows.
\greybox{
\sf \small
I am a robot that can pick and give objects to people. \\

task: give a person wearing a blue shirt a pepsi can.\\
scene: objects = [pepsi can, coca cola can, apple, banana]\\
scene: people = [person wearing blue shirt, person wearing black shirt]\\
robot action: robot.pick\_and\_give(pepsi can, person wearing blue shirt)\\
robot action: done()}
\greybox{
\sf \small

task: give a person wearing ligt colored shirt a monster can.\\
scene: objects = [pepsi can, coca cola can, monster can, lime]\\
scene: people = [person wearing yellow shirt, person wearing black shirt]\\
robot action: robot.pick\_and\_give(monster can, person wearing yellow shirt)\\
robot action: done()\\

task: give a person wearing a white shirt a coca cola can.\\
scene: objects = [starbucks can, coca cola can, monster can, lime]\\
scene: people = [person wearing red shirt, person wearing white shirt]\\
robot action: robot.pick\_and\_give(coca cola can, person wearing white shirt)\\
robot action: done()\\

task: give a person wearing black shirt a monster can.\\
scene: objects = [pepsi can, coca cola can, starbucks can, lime, apple]\\
scene: people = [person wearing green shirt, person wearing black shirt]\\
robot action: robot.pick\_and\_give(monster can, person wearing black shirt)\\
robot action: done()\\

task: considering the ambiguity of the goal, \orange{goal here} \\
scene: objects = [\orange{objects here}]\\
scene: people = [\orange{people here}]\\
}
The prompts used for classification and disambiguation are as follows. This is appended to the previous prompts.
\greybox{
\sf \small
robot thought: I am a robot that can only give the object to humans. Can I \orange{goal here}? \\
robot thought: \highlight{bpink}{\capability Generated Feasibility}
\\
Robot thought: this code is uncertain because \highlight{bubbles}{\reasoning Generated Reason} \\
Robot thought: what can I ask to user? \\
Question: Please \highlight{champagne}{\question Generated Question} \\
Answer: \orange{User's feedback} \\
robot thought: continue the previous task based on the question and answer
}
\subsection{Fewshot Prompts}
\paragraph{Pick and place Environment} Fewshot prompts with Inner Monologue style \cite{22_huang}. 
\greybox{\sf \small
task: move all the blocks to the top left corner. \\
scene: objects = [red block, yellow block, blue block, green bowl] \\
robot action: robot.pick\_and\_place(blue block, top left corner) \\
robot action: robot.pick\_and\_place(red block, top left corner) \\
robot action: robot.pick\_and\_place(yellow block, top left corner) \\
robot action: done()}
\greybox{
\sf \small

task: put the yellow one the green thing. \\
scene: objects = [red block, yellow block, blue block, green bowl] \\
robot action: robot.pick\_and\_place(yellow block, green bowl)
robot action: done() \\

task: move the light colored block to the middle. \\
scene: objects = [yellow block, blue block, red block]\\
robot action: robot.pick\_and\_place(yellow block, middle)\\
robot action: done()\\

task: stack all the blocks.\\
scene: objects = [blue block, green bowl, red block, yellow bowl, green block]\\
Robot action: robot.pick\_and\_place(green block, blue block)\\
Robot thought: This code is uncertain because the robot don’t know which block to pick up first\\
Robot thought: What can I ask to the user?\\
Question: which block should I pick up first?\\
Answer: green block\\
robot action: robot.pick\_and\_place(green block, blue block)\\
robot action: robot.pick\_and\_place(red block, green block)\\
done()\\

task: group the blue objects together.\\
scene: objects = [red block, blue block, green bowl, blue bowl, yellow block, green block]\\
robot action: robot.pick\_and\_place(blue block, blue bowl)\\
robot action: done()\\

task: put all blocks in the green bowl.\\
scene: objects = [red block, blue block, green bowl, blue bowl, yellow block]\\
robot action: robot.pick\_and\_place(red block, green bowl)\\
robot action: robot.pick\_and\_place(blue block, green bowl)\\
robot action: robot.pick\_and\_place(yellow block, green bowl)\\
robot action: done()\\
}
\paragraph{SaGC Dataset}
Prompts for cooking robots are as follows. 
\greybox{
\sf \small
I am a cooking robot. Your possible action is grab, heat and plate. \\
objects: [bread, pan, table, water] \\
goal: cook me toast \\
robot action: robot.grab(bread) \\
robot action: robot.heat(bread) \\
robot action: robot.plate(bread)  \\
robot action: robot.done()\\

objects: [table, watter, coffemachine] \\
goal: give a person wearing black shirt a coffee \\
robot action: robot.grab(coffee) \\
robot action: robot.pour(coffee)  \\
robot action: robot.done()\\

objects: [bread, pan, bacon, watter, table] \\
goal: give me a sandwich \\
robot action: robot.grab(bread) \\
robot action: robot.heat(bread) \\
robot action: robot.grab(bacon) \\
robot action: robot.heat(bacon) \\
robot action: robot.plate(bread) }
\greybox{
\sf \small
robot action: robot.plate(bacon) \\
robot action: robot.done()\\

objects: [water, table, pan, bread, 'coffee] \\
goal: I am thirsty\\
robot thought: I am not sure what to do grab first\\
question: What should I do first?\\
answer: grab water\\
robot action: robot.grab(water)\\
robot action: robot.pour(water)\\
robot action: robot.done()\\

objects: [water, table, pan, bread, coffee] \\
goal: wipe here \\
robot thought: I can not do this task \\
question: provide more information about what to do \\
answer: I can not do this task \\
robot action: robot.done()\\

From this, predict the next action by considering the role of the robot and the ambiguity of the goal \\
objects: [\orange{Objects here}]\\
goals: \orange{Goal Here}
}
Prompts for cleaning robots are as follows. 
\greybox{
\sf \small
I am a cleaning robot. Your possible action is clean and wipe. \\
scene: [kitchen, living room, bedroom, bathroom] \\
objects: [table, pan, bread] \\
goal: clean the kitchen\\
robot action: robot.clean(kitchen)\\
robot action: robot.done()\\

scene: [kitchen, living room, bedroom]\\
objects: [table, pan, bread] \\
goal: clean bedroom\\
robot action: robot.clean(bedroom)\\
robot action: robot.done()\\

scene: [kitchen, living room, bedroom] \\
objects: [table, pan, bread] \\
goal: clean table\\
robot action: robot.wipe(table)\\
robot action: robot.done()\\

scene: [kitchen, living room, bedroom] \\
objects: [table, pan, desk, bed] \\
goal: clean here\\
robot thought: this is uncertain because I am not sure what is here\\
question: what is here\\
answer: desk\\
robot action: robot.wipe(desk)\\
robot action: robot.done()\\

objects: [water, table, pan, bread, coffee]\\
goal: heat bread\\
robot thought: I can not do this task\\
question: provide more information about what to do\\
answer: I can not do this task\\
robot action: robot.done()}
\greybox{
\sf \small
From this, predict the next action by considering the role of the robot and the ambiguity of the goal \\
scene: [\orange{Scene here}]\\
objects: [\orange{Objects here}]\\
goals: \orange{Goal Here}
}
Prompts for massage robots are as follows. 
\greybox{
\sf \small
I am a massage robot. Your possible action is go\_to and give\_massage. \\

people: [person wearing black shirt, person wearing yellow shirt, person wearing red shirt] \\
goal: Give a massage to a person wearing black shirt\\
robot action: robot.go\_to(person wearing black shirt)\\
robot action: robot.give\_massage(person wearing black shirt)\\
robot action: robot.done()\\

poeple: [person wearing green shirt, person wearing green shirt, person wearing blue shirt]\\
goal: Give a massage to a person wearing green shirt\\
robot action: robot.go\_to(person wearing green shirt)\\
robot action: robot.give\_massage(person wearing green shirt)\\
robot action: robot.done()\\

poeple: [person wearing green shirt, person wearing green shirt, person wearing blue shirt]\\
goal: A person wearing green shirt needs a massage\\
robot action: robot.go\_to(person wearing green shirt)\\
robot action: robot.give\_massage(person wearing green shirt)\\
robot action: robot.done()\\

people: [person wearing black shirt, person wearing yellow shirt, person wearing red shirt] \\
goal: Someone needs a massage\\
robot thought: this is uncertain because I am not sure who needs a massage\\
question: who needs a massage\\
answer: person wearing yellow shirt\\
robot action: robot.go\_to(person wearing yellow shirt)\\
robot action: robot.give\_massage(person wearing yellow shirt)\\
robot action: robot.done()\\

people: [person wearing black shirt, person wearing yellow shirt, person wearing red shirt] \\
goal: heat bread\\
robot thought: I can not do this task\\
question: provide more information about what to do\\
answer: I can not do this task\\
robot action: robot.done()\\
        
From this, predict the next action by considering the role of the robot and the ambiguity of the goal \\
people: [\orange{people here}] \\
goals: \orange{Goal Here} \\
}
\paragraph{Real-world Environment} The prompts used are as follows.
\greybox{
\sf \small
I am a robot that can pick and give objects to people. \\

task: give a person wearing a blue shirt a pepsi can.\\
scene: objects = [pepsi can, coca cola can, apple, banana]\\
scene: people = [person wearing blue shirt, person wearing black shirt]\\
robot action: robot.pick\_and\_give(pepsi can, person wearing blue shirt)\\
robot action: done()\\

task: give a person wearing ligt colored shirt a monster can.\\
scene: objects = [pepsi can, coca cola can, monster can, lime]\\
scene: people = [person wearing yellow shirt, person wearing black shirt]\\
robot action: robot.pick\_and\_give(monster can, person wearing yellow shirt)\\
robot action: done()\\

task: give a person wearing a white shirt a coca cola can.\\
scene: objects = [starbucks can, coca cola can, monster can, lime]\\
scene: people = [person wearing red shirt, person wearing white shirt]\\
robot action: robot.pick\_and\_give(coca cola can, person wearing white shirt)\\
robot action: done()\\

task: give a person wearing light colored shirt a monster can.\\
scene: objects = [pepsi can, coca cola can, monser can, lime]\\
scene: people = [person wearing yellow shirt, person wearing white shirt]\\
robot thought: This is code is uncertain because I don't know whom to give.\\
robot thought: What can I ask to the user?\\
question: Which person do you want to give monster can?\\
answer: person wearing yellow shirt \\
robot action: robot.pick\_and\_give(monster can, person wearing yellow shirt)\\
robot action: done()\\

task: person yellow shirts wants a handshakes.\\
scene: objects = [pepsi can, coca cola can, monser can, lime]\\
scene: people = [person wearing yellow shirt, person wearing black shirt]\\
robot thought: I can not do this task.\\
robot action: done()  \\

task: considering the ambiguity of the goal, \orange{goal here} \\
scene: objects = [\orange{objects here}]
scene: people = [\orange{people here}]\\
}

\subsection{Real-world Experiments Environment and Demonstrations}
\begin{figure}[!h]
    \centering
    \includegraphics[width=0.5\textwidth]{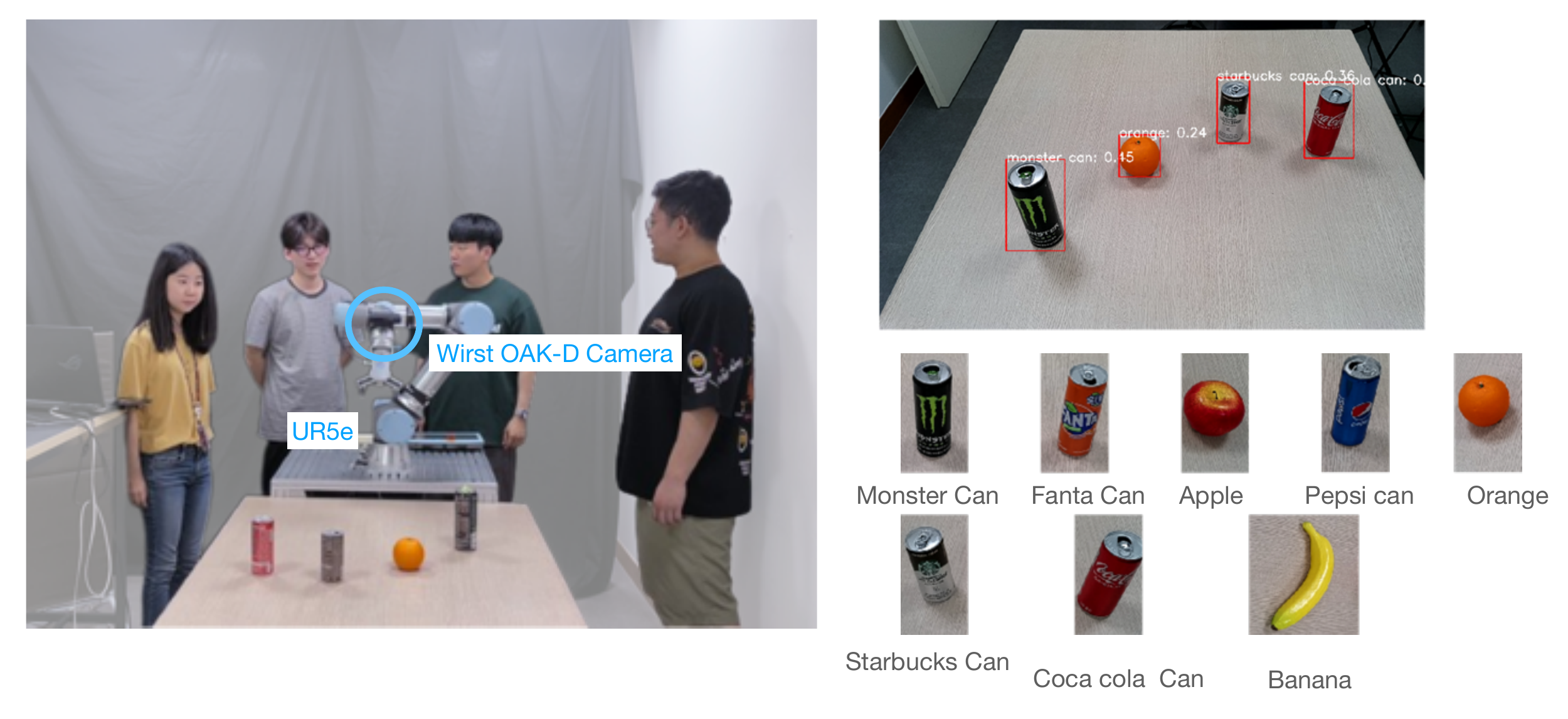}
    \caption{Real-world Experiment Setting}
    \label{fig:realworld_setup}
\end{figure}

\paragraph{Environments}
The setup consists of a UR5e robot arm equipped with a wrist-mounted OpenCV OAKD camera overlooking a workspace of the tabletop. We run our method via Robot Operating System (ROS) and the rate of publishing the joint position is set at 500HZ which is a default setting for a UR5e manipulator.
In the table scene, we randomly place objects on the table and shuffle them throughout the experiments. The setup is shown in Figure
\ref{fig:realworld_setup}.
\begin{figure*}[!h]
    \centering
    \includegraphics[width=0.99\textwidth]{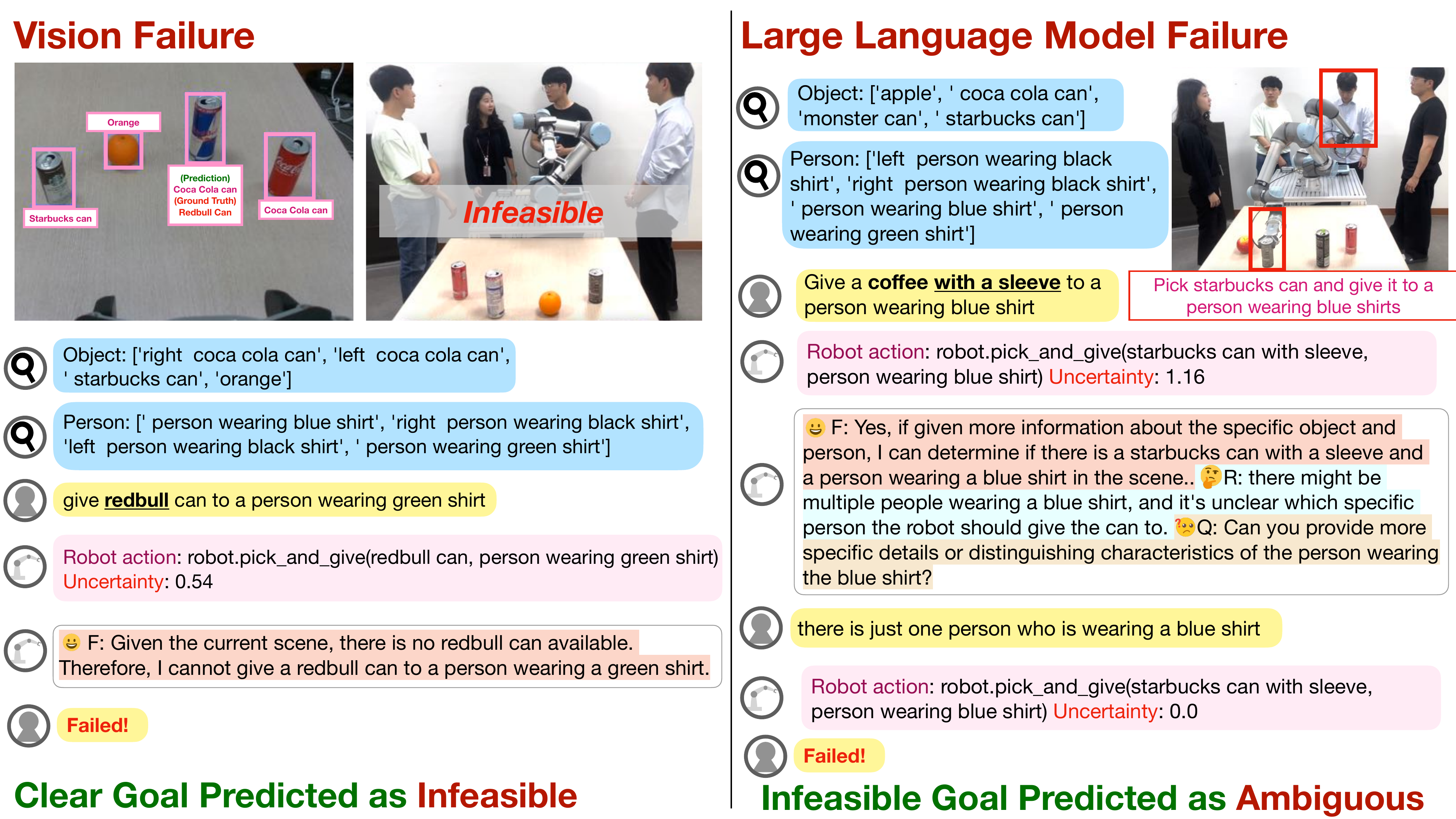}
    \caption{Failure Cases in Real-world Environment}
    \label{fig:failure_rw}
\end{figure*}

We investigate three tasks considering goal information: (i) Certain cases involve the explicit provision of information regarding the location or name of the target object. (ii) Ambiguous cases, in contrast, are characterized by the absence of detailed information about the target object. (iii) infeasible cases arise when objects or actions are encountered that have not been previously observed within the given context prompts. The goals are illustrated in Table \ref{tab:realword_tasks}. 
\begin{table}[!h]
    \centering
    \begin{tabular}{|c|c|}
        \hline
         Categories & Tasks\\
        \hline
        Clear & give [x] to [y]\\
        \hline
        \multirow{3}{*}{Ambiguous} & give [x] to someone \\
        & give something to drink to [y] \\
& give something to drink to someone \\ 
        \hline
        \multirow{2}{*}{Infeasible}
        & Wipe the desk \\
        & Smash the [x] \\
        & put [x] on the ground \\
        \hline
    \end{tabular}
    \caption{Task explanation for the real-world experiment.}
    \label{tab:realword_tasks}
\end{table}
\paragraph{User Guidelines}
A high-level goal was predefined during the experiment, as shown in Table V in the manuscript. There were four participants in each scene, and they were asked to answer the question from the robot during the disambiguation process. 
The instructions given to the participants are as follows:

\greybox{ 
\sf \small Given the goal, the robot may ask you some questions if it is 
uncertain about what to do. If the robot asks for the information, please 
answer it via the interface and as detailed as possible, letting the robot be certain about the task. \\

An example of the response is as follows: \\
Scene: [coca cola can, pepsi can, apple, lemon]\\
Q. Please specify what kind of drink you would like. \\
A. Coca cola can
}

\paragraph{Failure Cases}
The failure can arise due to both VLM and LLM. We have shown the failure cases in Figure \ref{fig:failure_rw}. On the vision failure part, the VLM fails to detect the Red Bull can, leading to predict \texttt{"give red bull can to a person wearing green shirt"} as in feasible command. In the LLM Failure part, LLM failed to understand that \texttt{coffee with sleeve} is infeasible in the scene and picks the Starbucks can instead. 

\paragraph{Detailed Results}
The detailed results on disambiguation in the real world are shown in Table \ref{tab:realworld_disambi}. We report the success rate of both before and after the interaction, along with a portion of question generation for each label. 
\begin{table*}[!h]
\centering
    \begin{tabular}{|c|c||c|c|c||c|c|c|}
        \hline
         Model &Method & Before & After & Gap & C. Por. & A. Por. & I. Por. \\
        \hline
         \multirow{3}{*}{GPT3.5} & IM \cite{22_huang}& 0.38 & 0.55 & 0.17  & 0.83 & 0.83 & 0.17 \\
         & CLAM\dag \cite{22_kuhn} & 0.33 & 0.61 & 0.28 & 0.66 & 1.0 & 0.33  \\
         & Ours & 0.22 & 0.55 & 0.33 & 0.33 & 1.0 & 0.33 \\
        \hline
        \multirow{3}{*}{IntructGPT} & IM \cite{22_huang} & 0.27 & 0.44 & 0.17 & 0.33 & 0.66 & 0.0  \\
        & CLAM\dag \cite{22_kuhn} & 0.33 & 0.39 & 0.06 & 0.33 & 0.33 & 0.16 \\
        & Ours & 0.38 & 0.67 & 0.28 & 0.17 & 0.83 & 0.0\\
        \hline
    \end{tabular}
    \caption{Success Rates on real-world environment. C. Por., A. Por., I. Por. denotes a portion of question generation on clear, ambiguous, and infeasible commands respectively. }
    \label{tab:realworld_disambi}
\end{table*}

\begin{figure*}[!h]
    \centering
    \includegraphics[width=0.8\textwidth]{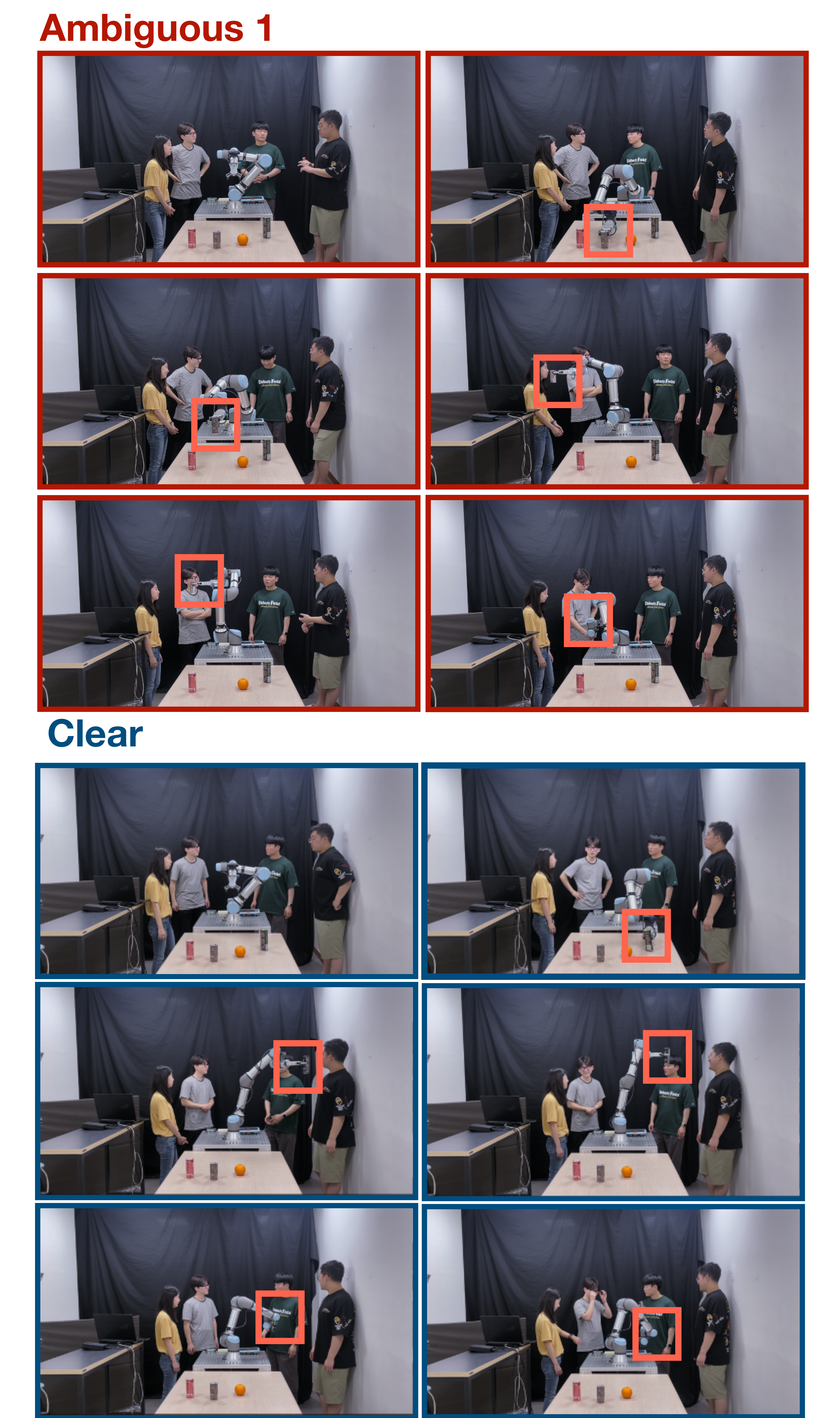}
    \caption{Snapshot1}
    \label{fig:realworld_snapshot1}
\end{figure*}
\begin{figure*}[!h]
    \centering
    \includegraphics[width=0.8\textwidth]{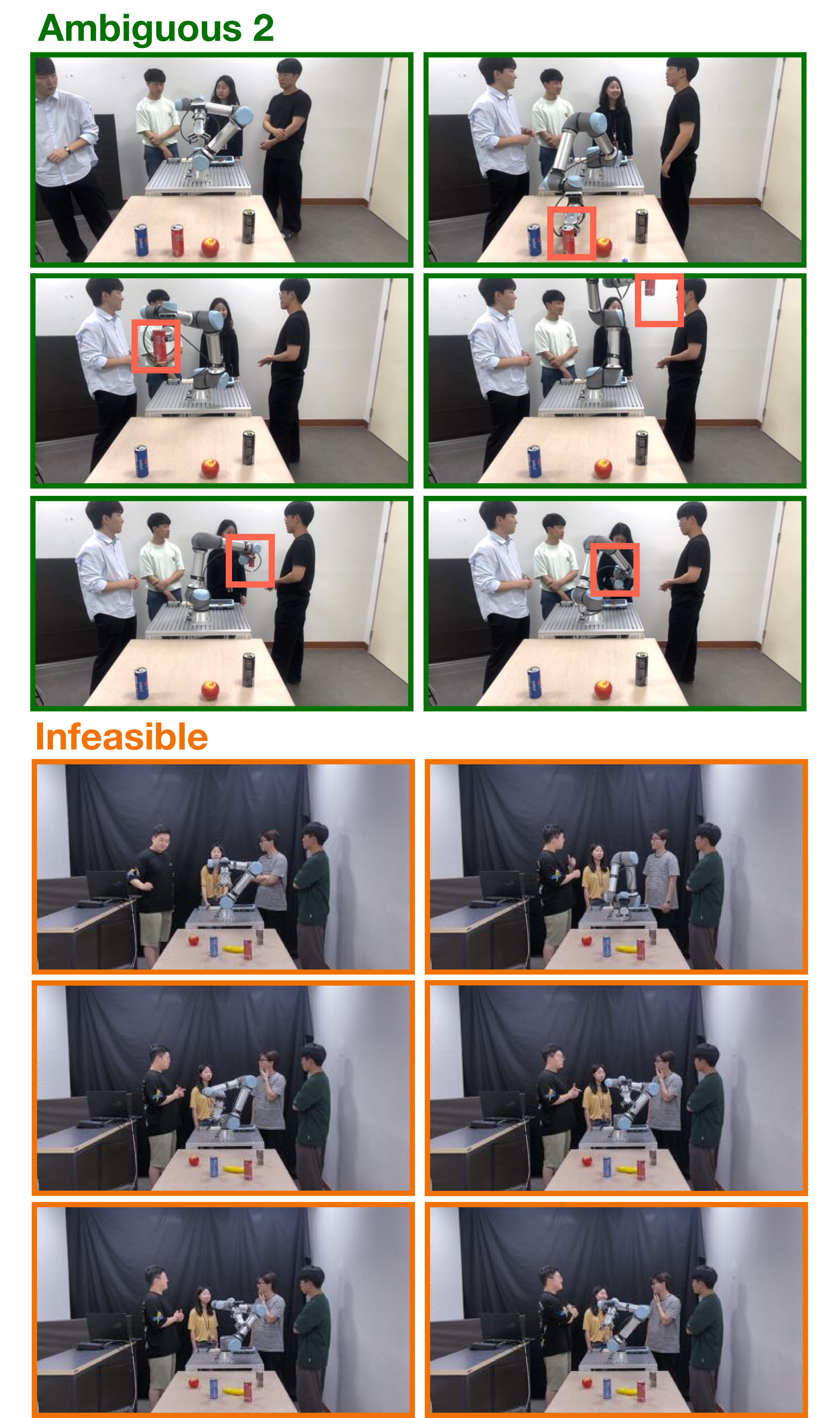}
    \caption{Snapshot2}
    \label{fig:realworld_snapshot2}
\end{figure*}

The snapshots for real-world demonstrations are shown in Figure \ref{fig:realworld_snapshot1} and \ref{fig:realworld_snapshot2}.

\subsection{Additional Results on SaGC Dataset}
\begin{figure*}[!h]
    \centering
    \includegraphics[width=.85\textwidth]{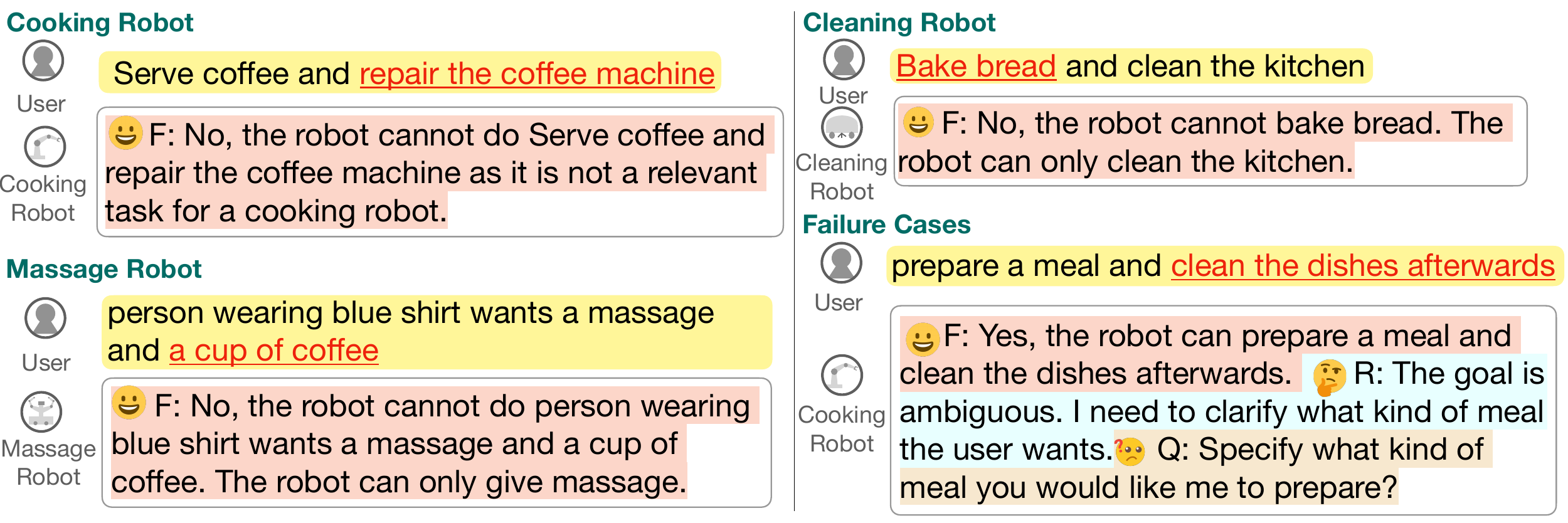}
    \caption{Generated output on the additional samples}
    \label{fig:sample}
\end{figure*}

The dataset contains partially feasible goals, such as \texttt{"bake buns and repair the desk"} for a cooking robot, \texttt{"cook breakfast and clean the kitchen} for a cleaning room, and \texttt{"make coffee and give a massage to the person wearing a gray shirt"} for massage robot. For additional $216$ data, we have validated the proposed method and observed that the proposed method on \texttt{text-davinci-003} model has the $0.87$ accuracy in those samples. Although the large-language model has hallucination issues, ensuring the model to be aware of their capabilities and roles helps in reducing those issues. The generated outputs in such samples are illustrated in Figure \ref{fig:sample}, with the failure case. 

\newpage

\end{document}